\documentclass[sigconf,bookmarksnumbered,unicode,pagebackref,breaklinks,colorlinks,allcolors=cvprblue]{acmart}

\usepackage{arydshln}
\usepackage{adjustbox}
\usepackage{amsmath}
\usepackage{makecell}
\usepackage{pifont}
\usepackage{url}            
\usepackage{booktabs}       
\usepackage{colortbl}       
\usepackage{amsfonts}       
\usepackage{nicefrac}       
\usepackage{microtype}      
\usepackage{xcolor}         
\usepackage{xcolor}
\usepackage{enumitem}
\usepackage{wrapfig}
\usepackage{algorithm}
\usepackage{multirow}
\usepackage{nccmath}
\usepackage{algpseudocode}
\usepackage{amsmath}

\newcommand{\best}[1]{\textbf{#1}}
\newcommand{\second}[1]{\underline{#1}}
\newcommand{\third}[1]{\textit{#1}}
\definecolor{cvprblue}{rgb}{0.21,0.49,0.74}
\definecolor{lightblue}{rgb}{0.796, 0.894, 0.9808}
\definecolor{naturegreen}{RGB}{0,102,85}

\AtBeginDocument{%
  }

\setcopyright{acmlicensed}
\copyrightyear{2018}
\acmYear{2018}
\acmDOI{XXXXXXX.XXXXXXX}
\acmConference[Conference acronym 'XX]{Make sure to enter the correct
  conference title from your rights confirmation email}{June 03--05,
  2018}{Woodstock, NY}
\acmISBN{978-1-4503-XXXX-X/2018/06}
\renewcommand\footnotetextcopyrightpermission[1]{} 

\setcopyright{none}

\begin{document}

\title{Rethinking Layer-Wise Information Allocation for Vision Foundation Model Adaptation}


\author{Yuqi Li}
\authornote{Both authors contributed equally to this work.}
\affiliation{%
  \institution{City College of New York, CUNY}
  \city{New York}
  \state{NY}
  \country{USA}
}
\email{yli@ccny.cuny.edu}

\author{Xi Xiao}
\authornotemark[1]
\affiliation{%
  \institution{University of Alabama at Birmingham}
  \city{Birmingham}
  \state{AL}
  \country{USA}
}
\email{xxiao@uab.edu}

\author{Yunbei Zhang}
\affiliation{%
  \institution{Tulane University}
  \city{New Orleans}
  \state{LA}
  \country{USA}
}
\email{yzhang@tulane.edu}

\author{Lin Zhao}
\affiliation{%
  \institution{Northeastern University}
  \city{Boston}
  \state{MA}
  \country{USA}
}
\email{l.zhao@northeastern.edu}

\author{Yu Li}
\affiliation{%
  \institution{George Washington University}
  \city{Washington}
  \state{DC}
  \country{USA}
}
\email{yli@gwu.edu}

\author{Aiden Zhao}
\affiliation{%
  \institution{Capital One}
  \city{McLean}
  \state{VA}
  \country{USA}
}
\email{aiden.zhao@capitalone.com}

\author{Tianyang Wang}
\affiliation{%
  \institution{University of Alabama at Birmingham}
  \city{Birmingham}
  \state{AL}
  \country{USA}
}
\email{twang@uab.edu}

\author{Hao Xu}
\affiliation{%
  \institution{Harvard University}
  \city{Cambridge}
  \state{MA}
  \country{USA}
}
\email{hxu@harvard.edu}

\author{Yingli Tian}
\authornote{Corresponding author.}
\affiliation{%
  \institution{City College of New York, CUNY}
  \city{New York}
  \state{NY}
  \country{USA}
}
\email{ytian@ccny.cuny.edu}

\renewcommand{\shortauthors}{Li et al.}

\begin{abstract}
Vision foundation models are increasingly reused as frozen backbones for downstream visual recognition, making parameter-efficient adaptation a central problem. Prompt-based adaptation, including Visual Prompt Tuning (VPT), provides a lightweight way to specialize these models, but its layer-wise behavior remains poorly understood: performance is sensitive to prompt depth, placement, and task distribution, and gains on standard in-domain benchmarks do not always translate into robust generalization. We argue that this limitation is not solely an optimization issue, but a layer-wise information allocation issue: existing prompt-based methods lack principled control over what prompt-conditioned representations should preserve, suppress, and propagate across depth. Inspired by the Information Bottleneck principle, we introduce \textbf{P}rompted \textbf{I}nformation \textbf{B}ottlenecks (\textbf{PIB}), a framework that regularizes layer-wise compression-sufficiency trade-offs and promotes a more coherent cross-layer information path. The key idea is that effective adaptation should be minimal yet sufficient, retaining task-relevant local evidence in earlier layers while progressively discarding nuisance factors and redundant details in deeper layers. Extensive experiments show that PIB achieves strong performance across 34 datasets, reaching 92.1\% on FGVC, 93.01\% on HTA, and 77.33\% on VTAB-1k, while tuning only 0.35\% parameters on average across the main settings. Beyond benchmark accuracy, PIB helps explain the non-monotonic behavior of prompt capacity scaling, reduces shortcut reliance, and improves robustness under distribution shift and fine-grained recognition settings. These results position PIB as both a practical method and an information-allocation perspective for adapting frozen vision foundation models. Our codes are available at \url{https://github.com/itsnotacie/MM-26-PIB}
\end{abstract}

\ccsdesc[500]{Information systems~Multimedia information systems}

\keywords{Information Bottleneck, PEFT, Vision Transformers}

\received{20 February 2007}
\received[revised]{12 March 2009}
\received[accepted]{5 June 2009}

\maketitle

\section{Introduction}

\begin{figure}[t]
    \centering
    \includegraphics[width=\linewidth]{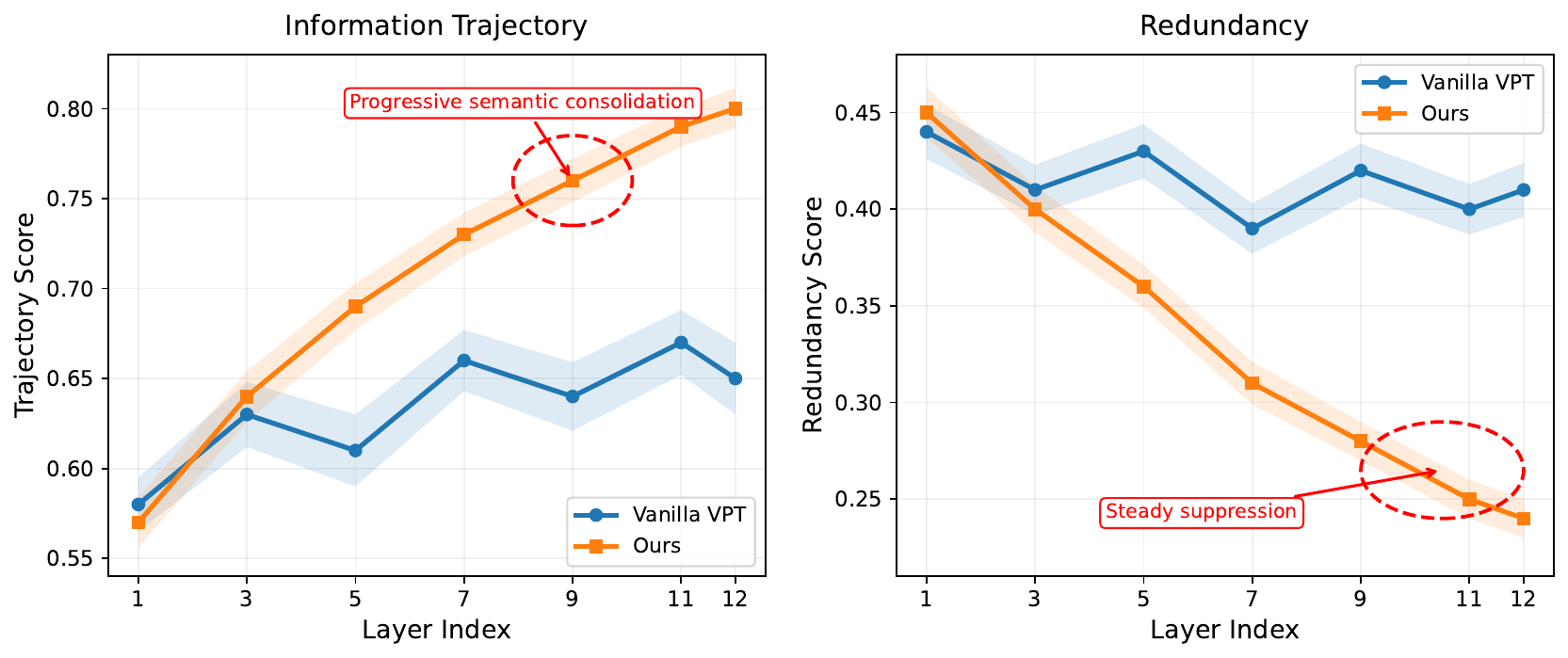}
    \caption{\textbf{A motivating diagnosis of layer-wise prompt behavior.}
    Compared with standard prompt tuning, our proposed PIB exhibits a more coherent information trajectory and a steadier reduction of redundancy across transformer depth. This trend suggests that a key limitation of prompt-based VFM adaptation is not merely insufficient prompt capacity, but the lack of principled control over layer-wise information allocation.}
    \label{fig:intro_diag}
\end{figure}

Large-scale vision foundation models (VFMs), especially Vision Transformers (ViTs), have become standard backbones for modern visual recognition and multimodal systems~\cite{dosovitskiy2021image,radford2021learning}. As these models are increasingly reused as frozen components, downstream adaptation has become a central challenge: practitioners care not only about transfer accuracy, but also about how to specialize pretrained models under constraints on compute, memory, and labeled data. In this context, parameter-efficient adaptation has emerged as a practical alternative to full fine-tuning.

Prompt-based adaptation is one representative family of such methods. Visual Prompt Tuning (VPT)~\cite{jia2022vpt}, for example, adapts a frozen transformer through a small set of learnable prompt tokens rather than updating the full backbone~\cite{xiao2025promptb}. This design is simple, modular, and compatible with a wide range of pretrained architectures. Recent work has extended prompt tuning through deeper prompt insertion, adaptive prompt generation, structured prompt design, and optimization-oriented refinements~\cite{han2023e2vpt,wang2024revisiting,zeng2024visual,xiao2025visual,wang2026visual}. However, despite this rapid progress, prompt-based VFM adaptation remains largely empirical: most existing methods focus on \emph{how} to design better prompts, while much less is understood about \emph{what} prompt-conditioned representations actually preserve, suppress, and propagate inside a frozen vision transformer.

Our starting point is a set of empirical observations that are difficult to explain under this design-centric view. Increasing prompt depth, prompt length, or insertion range does not reliably improve transfer performance. In some cases, stronger prompt parameterization helps on standard in-domain benchmarks but becomes unstable under fine-grained recognition, low-shot transfer, or distribution shift. We also observe that large prompt-induced changes in intermediate representations do not always lead to stronger semantic discrimination. Figure~\ref{fig:intro_diag} provides a representative diagnosis of this phenomenon: compared with PIB, standard deep prompting exhibits a less coherent cross-layer trajectory and weaker redundancy suppression. Taken together, these observations suggest that the weakness of prompt-based adaptation cannot be fully explained by prompt capacity or training stability alone.

Recent work has begun to move beyond purely structural prompt design. Prior analysis shows that the gap between prompt tuning and full fine-tuning cannot be reduced to simple optimization or overfitting arguments~\cite{han2024facing}, while other work attributes prompt-tuning limitations to unstable cross-layer prompt dynamics~\cite{xiao2025visual}. Our view is complementary but different. We argue that these phenomena point to a common failure mode: \textbf{\emph{existing prompt-based adaptation methods lack principled control over layer-wise information allocation.}} As a result, prompts may over-preserve nuisance factors such as texture, background, or dataset-specific statistics, or suppress task-relevant local evidence before robust semantics emerge.

This paper studies prompt-based VFM adaptation from the perspective of layer-wise information allocation. Rather than asking only how to make prompts more expressive, we ask: \textbf{\emph{how can we characterize and regulate the layer-wise information flow induced by prompts so that frozen-backbone adaptation learns minimal-yet-sufficient task representations?}} To answer this question, we draw inspiration from the Information Bottleneck (IB) principle~\cite{tishby2000information,tishby2015deep}. Instead of viewing prompts as auxiliary tokens appended to a frozen transformer, we interpret them as mechanisms that reshape the information path of intermediate representations across depth. Under this view, effective adaptation should follow a \emph{minimal-yet-sufficient} trajectory: early layers retain local evidence necessary for discrimination, while deeper layers progressively suppress nuisance factors and redundant detail.

Based on this formulation, we develop a training framework that regularizes layer-wise compression-sufficiency trade-offs and encourages more coherent cross-layer prompt behavior. Our empirical study is designed to answer three questions: \ding{182} Can this view explain the non-monotonic and sometimes fragile behavior of prompt capacity scaling? \ding{183} Can explicit information regularization reduce shortcut reliance and improve robustness beyond standard in-domain evaluation? \ding{184} Does a prompted information bottleneck lead to more interpretable layer-wise prompt dynamics? Through motivating analyses, controlled comparisons, and downstream experiments, we show that this perspective provides both a clearer explanation of prompt-based adaptation failure modes and a practical route to improving frozen VFM adaptation.

Our contributions are as follows:
\begin{itemize}
    \item We identify a previously underexplored failure mode in prompt-based VFM adaptation: its limitation arises not only from optimization dynamics, but also from unregulated layer-wise information allocation.
    \item We formulate prompt-conditioned frozen-backbone adaptation as a \emph{layer-wise Prompted Information Bottleneck (PIB) process}, providing an IB-inspired view in which effective prompts should follow a minimal-yet-sufficient information path across transformer depth.
    \item We develop a training framework that explicitly regularizes prompt-induced information allocation through complementary compression and sufficiency objectives together with cross-layer path constraints.
    \item We design empirical analyses that connect non-monotonic scaling, unstable generalization, shortcut sensitivity, and prompt dynamics to the proposed information-allocation perspective.
\end{itemize}

\section{Related Work}

\subsection{Parameter-Efficient Adaptation in Vision}

Parameter-efficient adaptation in vision has been explored through adapters, side branches, selective fine-tuning, low-rank updates, and prompt-based methods~\cite{dosovitskiy2021image,yosinski2014transferable,chen2020improved,zhang2020side,rebuffi2017learning,cai2020tinytl,hu2022lora,chen2022adaptformer,dong2023efficient,xu2025fakeshield,xu2024vision,xu2025dancefix,li2026comprehensive}. Prompt-based adaptation is especially attractive because it can specialize a frozen Vision Transformer by introducing only a small number of learnable tokens, with VPT serving as a representative instantiation~\cite{jia2022vpt}. Follow-up work has extended this family through deeper prompt insertion, adaptive prompt generation, structured prompting, and optimization-oriented refinements~\cite{jia2022vpt,han2023e2vpt,das2023learning,huang2023diversity,pei2024sa2vp,wang2024revisiting,zeng2024visual,jin2025lor,ren2025da-vpt,xiao2025visual,wang2026visual,xu2025quality,Xu_2025_CVPR}. Despite their differences, most methods are driven by the same question: how can prompts be made more expressive or easier to optimize? More recent studies begin to analyze prompt behavior itself, showing that prompt tuning may suffer from incomplete transfer or unstable cross-layer dynamics~\cite{han2024facing,xiao2025visual,wang2026visual}. Our work is related in spirit but different in focus. Rather than asking how prompts should be designed or stabilized, we ask what prompt-conditioned representations should preserve and discard across layers. This shifts the discussion from prompt structure or dynamics to layer-wise information allocation. Prompt-based adaptation has also moved beyond static, dataset-level token prompts. V$^2$APT generates input-dependent prompts through a variational latent model, while differentiable search can discover layer-specific prompt--token fusion rules~\cite{xiao2025v2apt,xiao2026layer}. In-context prompt tuning has been used to personalize large vision-language models from reference images~\cite{li2026personalize}, and dynamic prompts have also been explored for graph-based drug--target affinity prediction~\cite{xiao2024hgtdp}. These methods demonstrate the broad utility of content-adaptive prompting, but they do not study the layer-wise compression--sufficiency trajectory targeted by PIB. Efficiency in visual learning can also be pursued at the data and communication levels. Dataset-distillation methods use diffusion or coarse-to-fine autoregressive generators to construct compact, representative training sets~\cite{zhao2025taming,zhao2026hieramp}, while THInImg encodes long-form audio within an identity image for cross-modal talking-head reconstruction~\cite{zhao2024thinimg} and reasoning related~\cite{zhang-etal-2026-logical,zhang-etal-2026-semantic}. These directions reduce data or transmission costs and are complementary to our focus on parameter-efficient adaptation inside frozen VFMs.

\subsection{Information Bottleneck and Robustness}

Our work is also related to the Information Bottleneck (IB) literature and its deep-learning extensions~\cite{tishby2000information,tishby2015deep,saxe2019information,slonim2002information,alemi2017deep,chechik2003information,wu2020graph,goldfeld2020information,slonim1999agglomerative,zhang2026coupling,zhang2026intervensiminterventionawaresocialnetwork}. IB seeks representations that remove irrelevant input variability while retaining information necessary for the target task. While this perspective has influenced representation learning, robustness, and generalization analysis~\cite{chen2025temporal,vera2018collaborative,yuan2024dynamic,pan2021disentangled,wan2021multi,piran2020dual,zhang2025information,xiao2026not,elidan2005learning,federici2020learning,wu2020learnability}, most prior work studies fully trainable models rather than frozen-backbone adaptation. Our study is also connected to studies on shortcut learning and spurious visual cues, which show that strong in-domain performance may rely on unstable correlations such as texture or background~\cite{tang2021augmented,nuriel2022textadain,li2023whac,suhail2026shortcut,liu2019bow,wang2010fabric,chung2022shape,azad2021texture,jagadeesh2024texture,theodoridis2022trapped,hermann2020origins,zhou2025mitigating,suhail2026shortcut}. These observations are especially relevant to prompt-based frozen-backbone adaptation, where a small set of trainable prompts may amplify dataset-specific regularities instead of robust task semantics. However, prior work does not provide a prompt-specific framework for regulating this behavior across transformer layers. Our work fills this gap by formulating prompt-conditioned adaptation as an IB-inspired layer-wise prompted information bottleneck process.

\section{Method}

\begin{figure*}[t]
    \centering
    \includegraphics[width=\textwidth]{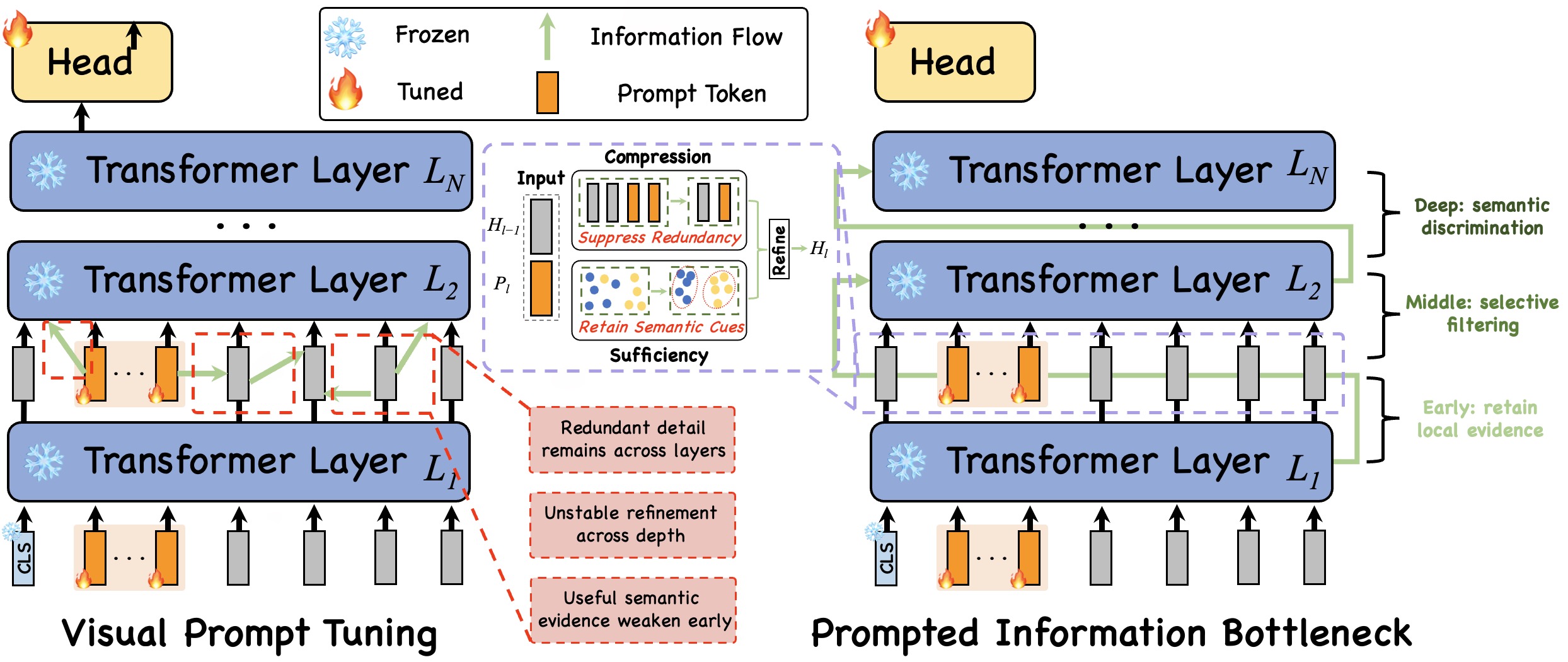}
    \caption{\textbf{Overview of our proposed Prompted Information Bottleneck (PIB).}
    \textbf{A}: prompt-based adaptation inserts trainable prompts into a frozen vision backbone without explicitly regulating what information should be preserved or suppressed, which may lead to redundant detail propagation, unstable cross-layer refinement, and early weakening of useful semantic cues.
    \textbf{B}: PIB introduces an IB-inspired layer-wise objective through complementary compression and sufficiency surrogates, while enforcing a minimal-sufficient path across depth from local evidence retention to semantic discrimination.
    \textbf{C}: PIB yields smoother redundancy reduction, stronger class separability, and a more ordered cross-layer path.}
    \label{fig:main_framework}
\end{figure*}

As illustrated in Figure~\ref{fig:main_framework}, we reinterpret prompt-based VFM adaptation as a layer-wise information allocation process and regularize it through complementary compression and sufficiency objectives together with a cross-layer minimal-sufficient path.

\subsection{Problem Setting}

We consider prompt-based adaptation of a pretrained vision foundation model, instantiated by a Vision Transformer (ViT) backbone \(F\) with \(L\) frozen transformer blocks. Given an input image \(x \in \mathcal{X}\) and label \(y \in \mathcal{Y}\), the patch embedding layer produces an initial token matrix \(H_0 \in \mathbb{R}^{N \times d}\), where \(N\) is the number of image tokens and \(d\) is the embedding dimension. A task head \(g(\cdot)\) produces the final prediction from the last-layer representation.

Following VPT-style adaptation, we introduce trainable prompt parameters \(P_1, \dots, P_L\), where \(P_l \in \mathbb{R}^{M_l \times d}\) denotes the prompt associated with layer \(l\). We keep the formulation general and only assume that prompts affect the layer-wise hidden states. The prompt-conditioned representation at layer \(l\) is written as
\[
H_l = f_l(H_{l-1}, P_l), \qquad l = 1, \dots, L,
\]
where \(f_l\) denotes the frozen \(l\)-th transformer block together with the prompt interaction mechanism. The final prediction is
\[
\hat{y} = g(H_L),
\]
and standard prompt tuning optimizes only the task loss
\[
\mathcal{L}_{\mathrm{task}} = \ell(\hat{y}, y).
\]

The limitation of this objective is that it specifies only the output target, but not what prompt-conditioned representations should preserve or suppress across layers. As a result, prompts may retain nuisance-heavy information for too long or discard task-relevant evidence too early. 

\subsection{IB-inspired Prompted Information Bottleneck}

We reinterpret prompt tuning as a process of layer-wise information allocation. Instead of viewing prompts as auxiliary tokens, we view them as lightweight control variables that reshape how information flows through a frozen transformer. Under this view, successful adaptation should preserve useful evidence in early layers while progressively filtering nuisance factors in deeper layers.

Our formulation is inspired by the Information Bottleneck (IB) principle~\cite{tishby2000information,tishby2015deep}, which seeks representations that compress irrelevant input variability while preserving task-relevant information. For a representation \(Z\), the classical IB objective is
\[
\min \; I(Z;X) - \beta I(Z;Y),
\]
where \(I(\cdot;\cdot)\) denotes mutual information. In our setting, we extend this idea to the layer-wise prompt-conditioned representation chain \(H_1,\dots,H_L\), leading to the idealized objective
\[
\min_{\{P_l\}} \sum_{l=1}^{L} \Big( I(H_l; X) - \beta_l I(H_l; Y) \Big).
\]

This objective captures our main hypothesis: prompt tuning should follow a minimal-yet-sufficient information path across depth. Since direct mutual information optimization is intractable in this setting, we instead introduce practical surrogates for compression and sufficiency.

\subsection{Layer-wise Compression and Sufficiency}

We decompose the prompted information bottleneck into two complementary terms:
\[
\mathcal{L}_{\mathrm{PIB}} = \sum_{l=1}^{L} \Big( \lambda_l \mathcal{L}^{(l)}_{\mathrm{comp}} + \gamma_l \mathcal{L}^{(l)}_{\mathrm{suff}} \Big),
\]
where \(\lambda_l\) and \(\gamma_l\) control the trade-off at layer \(l\).

\paragraph{Compression.}
The compression term discourages prompts from propagating excessive redundant or nuisance-heavy information. Since exact \(I(H_l;X)\) is unavailable, we adopt a redundancy-based surrogate. Let \(\tilde{H}_l \in \mathbb{R}^{N_l \times d}\) denote the row-wise normalized hidden representation at layer \(l\), and define the empirical correlation matrix
\[
C_l = \frac{1}{N_l} \tilde{H}_l^{\top} \tilde{H}_l.
\]
We then use the off-diagonal energy of \(C_l\) as the compression loss:
\[
\mathcal{L}^{(l)}_{\mathrm{comp}} = \| C_l - I_d \|_{F,\mathrm{off}}^2,
\]
where \(I_d\) is the identity matrix. Minimizing this term encourages less redundant layer-wise representations.

\paragraph{Sufficiency.}
Compression alone is not enough, since excessive compression may remove useful discriminative cues. We therefore add a sufficiency term based on class separability. Let \(h_{l,i} \in \mathbb{R}^{d}\) denote the pooled representation of sample \(i\) at layer \(l\), and let \(\mu_{l,c}\) be the corresponding class mean in a mini-batch. We define the within-class and between-class scatter as
\[
S^{(l)}_{\mathrm{intra}} = \sum_{c} \sum_{i \in \mathcal{B}_c} \| h_{l,i} - \mu_{l,c} \|_2^2,
\]
\[
S^{(l)}_{\mathrm{inter}} = \sum_{c \neq c'} \| \mu_{l,c} - \mu_{l,c'} \|_2^2.
\]
The sufficiency loss is then
\[
\mathcal{L}^{(l)}_{\mathrm{suff}} = \frac{S^{(l)}_{\mathrm{intra}}}{S^{(l)}_{\mathrm{inter}} + \epsilon},
\]
where \(\epsilon\) is a small constant. This encourages each layer to remain discriminative while being compressed.

Together, these two terms implement the desired minimal-sufficient trade-off in a tractable form: compression suppresses redundancy, while sufficiency preserves task-relevant structure. We do not treat these losses as exact mutual-information estimators; rather, they are optimization-friendly surrogates aligned with the IB-inspired objective above.

\subsection{Cross-layer Minimal-Sufficient Path}

Regulating each layer independently is still insufficient, since successful prompt tuning should also follow a coherent refinement path across depth. Early layers should remain relatively detail-preserving, while deeper layers should become progressively more selective.

To capture this behavior, we define a layer quality score
\[
q_l = \mathcal{L}^{(l)}_{\mathrm{comp}} + \alpha_l \mathcal{L}^{(l)}_{\mathrm{suff}},
\]
where a lower score indicates a better compression-sufficiency balance. We then regularize the cross-layer sequence through
\[
\mathcal{L}_{\mathrm{path}} = \sum_{l=1}^{L-1} \max \big( 0, q_{l+1} - q_l - \delta_l \big),
\]
where \(\delta_l \geq 0\) is a tolerance margin. This term penalizes abrupt deterioration in the layer-wise trade-off and encourages a smoother transition from local evidence preservation to higher-level semantic selectivity.

\subsection{Optional Layer-aware Routing}

Different layers need not share the same compression--sufficiency balance. We therefore introduce an optional learnable gate \(r_l \in [0,1]\) that assigns a layer-dependent weighting to the two objectives:
\[
\mathcal{L}_{\mathrm{PIB}}^{\mathrm{route}} =
\sum_{l=1}^{L} \Big( r_l \lambda_l \mathcal{L}^{(l)}_{\mathrm{comp}} + (1-r_l)\gamma_l \mathcal{L}^{(l)}_{\mathrm{suff}} \Big).
\]
A larger \(r_l\) places more weight on redundancy suppression, whereas a smaller value favors preservation of discriminative evidence. The gates are learned jointly with the prompts and serve as an optional refinement of the base formulation. Parameterization and implementation details are provided in Appendix~\ref{app:routing_impl}.

\subsection{Final Objective}

Our final objective combines the task loss with the proposed regularizers:
\[
\mathcal{L}_{\mathrm{total}} =
\mathcal{L}_{\mathrm{task}}
+ \mathcal{L}_{\mathrm{PIB}}
+ \eta \mathcal{L}_{\mathrm{path}},
\]
where \(\eta\) controls the strength of path regularization. The routed variant replaces \(\mathcal{L}_{\mathrm{PIB}}\) with the layer-aware objective defined in Appendix~\ref{app:routing_impl}.

During training, we optimize the prompt parameters, the task head, and optionally the routing variables, while keeping the pretrained transformer blocks frozen. The key difference from standard prompt tuning is therefore not a new prompt structure, but an explicit principle for regulating prompt behavior: prompts should preserve what the task needs, suppress what it does not, and do so progressively across layers.

\section{Experiments}

\begin{table*}[!htbp]
\centering
    \caption{
    \textbf{Comparison of fine-tuning methods with ViT-Base/16 backbone.} \textbf{Bold} and \underline{underlined} indicate best and second-best results. We report mean accuracy across datasets for each benchmark (FGVC: 5 datasets, HTA: 10 datasets, VTAB-1k: 19 datasets), for each VTAB-1k category (Natural, Specialized, and Structured). Please see Appendix for per-task results.
    Same for Table \ref{table:mae_moco}.}
\label{tab:full_comparison}

\scriptsize 

\begin{adjustbox}{width=1\textwidth,center}
\begin{tabular}{c||c|c|c|c|ccc|c} 
\Xhline{4\arrayrulewidth}
\rowcolor{gray!20}
ViT-Base/16~\cite{dosovitskiy2021image} & Tuned/Total &  & FGVC Mean & HTA Mean & \multicolumn{3}{c|}{VTAB-1k~\cite{zhai2019large}} & VTAB-1k Mean \\ 
\rowcolor{gray!20}
\rowcolor{gray!20}
(85.8M) & (\%) & \multirow{-2}{*}{Extra Params} & (\%) & (\%) & \textit{Natural} & \textit{Specialized} & \textit{Structured} & (\%) \\ 
\hline \hline
Full \textcolor{lightgray}{\scriptsize{[CVPR22]}}\cite{iofinova2022well} & 100.00 & — & 88.54 & 85.8 & 75.88 & 83.36 & 47.64 & 65.57 \\ 
\hline
Linear \textcolor{lightgray}{\scriptsize{[CVPR22]}}\cite{iofinova2022well} & 0.08 & — & 79.32 & 75.7 & 68.93 & 77.16 & 26.84 & 52.94 \\ 
Partial-1 \textcolor{lightgray}{\scriptsize{[NeurIPS14]}}\cite{yosinski2014transferable} & 8.34 & — & 82.63 & 80.8 & 69.44 & 78.53 & 34.17 & 56.52 \\ 
MLP-3 \textcolor{lightgray}{\scriptsize{[CVPR20]}}\cite{chen2020improved} & 1.44 & \checkmark & 79.80 & 78.5 & 67.80 & 72.83 & 30.62 & 53.21 \\ 
\hline
Sidetune \textcolor{lightgray}{\scriptsize{[ECCV20]}}\cite{zhang2020side} & 10.08 & — & 78.35 & 72.3 & 58.21 & 68.12 & 23.41 & 45.65 \\ 
Bias \textcolor{lightgray}{\scriptsize{[NeurIPS17]}}\cite{rebuffi2017learning} & 0.80 & — & 88.41 & 82.1 & 73.30 & 78.25 & 44.09 & 62.05 \\ 
Adapter \textcolor{lightgray}{\scriptsize{[NeurIPS20]}}\cite{cai2020tinytl} & 1.02 & \checkmark & 85.46 & 80.6 & 70.67 & 77.80 & 33.09 & 62.41 \\ 
LoRA \textcolor{lightgray}{\scriptsize{[ICLR22]}}\cite{hu2022lora} & — & \checkmark & 89.46 & 85.5 & 78.26 & 83.78 & 56.20 & 72.25 \\ 
AdaptFormer \textcolor{lightgray}{\scriptsize{[NeurIPS22]}}\cite{chen2022adaptformer} & — & \checkmark & — & — & 80.56 & 84.88 & 58.83 & 72.32 \\ 
ARC$_{\text{att}}$ \textcolor{lightgray}{\scriptsize{[NeurIPS23]}}\cite{dong2023efficient} & — & \checkmark & 89.12 & 89.0 & 80.41 & 85.55 & 58.38 & 72.32 \\ 
\hline
VPT-S \textcolor{lightgray}{\scriptsize{[ECCV22]}}\cite{jia2022vpt} & 0.16 & \checkmark & 84.62 & 85.5 & 76.81 & 79.66 & 46.98 & 64.85 \\ 
VPT-D \textcolor{lightgray}{\scriptsize{[ECCV22]}}\cite{jia2022vpt} & 0.73 & \checkmark & 89.11 & 85.5 & 78.48 & 82.43 & 54.98 & 69.43 \\
E2VPT \textcolor{lightgray}{\scriptsize{[ICCV23]}}\cite{han2023e2vpt} & 0.39 & \checkmark & 89.22 & 88.5 & 80.01 & 84.43 & 57.39 & 71.42 \\ 
EXPRES \textcolor{lightgray}{\scriptsize{[CVPR23]}}\cite{das2023learning} & — & \checkmark & — & — & 79.69 & 84.03 & 54.99 & 70.02 \\ 
DAM-VP \textcolor{lightgray}{\scriptsize{[CVPR23]}}\cite{huang2023diversity} & — & \checkmark & — & 88.5 & — & — & — & — \\ 
SA\textsuperscript{2}VP \textcolor{lightgray}{\scriptsize{[AAAI24]}}\cite{pei2024sa2vp} & 0.81 & \checkmark & \underline{90.08} & 91.5 & 80.97 & \underline{85.73} & 60.80 & 75.83 \\
SPT \textcolor{lightgray}{\scriptsize{[ICML24]}}\cite{wang2024revisiting} & — & \checkmark & 90.10 & — &  81.44 &  83.65 & 52.86 & 72.65 \\
VFPT \textcolor{lightgray}{\scriptsize{[NeurIPS24]}}\cite{zeng2024visual} & 0.66 & \checkmark & 89.24 & — & 81.35 & 84.93 & 60.19 & 75.49 \\
LoR-VP \textcolor{lightgray}{\scriptsize{[ICLR25]}}\cite{jin2025lor} & — & \checkmark & 91.22 & — & 80.25 & 85.12 & 58.71 & 74.69 \\
DA-VPT \textcolor{lightgray}{\scriptsize{[CVPR25]}}\cite{ren2025da-vpt} & — & \checkmark & 89.32 & — & 79.91 & 83.16 & 60.01 & 74.36 \\
ViaPT \textcolor{lightgray}{\scriptsize{[ACMMM25]}}\cite{xiao2025visual} & 0.66 & \checkmark & \underline{91.4} & \underline{92.20} & \underline{82.62} & 85.22 & \underline{61.25} & \underline{76.36} \\
\rowcolor{cvprblue!15} 
\textbf{Ours} & 0.51 & \checkmark & \textbf{92.1} & \textbf{93.01} & \textbf{82.98} & \textbf{85.83} & \textbf{63.18} & \textbf{77.33} \\ 
\Xhline{4\arrayrulewidth}
\end{tabular}
\end{adjustbox}
\label{tab:vit}
\end{table*}

\begin{table*}[!ht]
\caption{\textbf{Comparison of the fine-tuning methods under different pretraining paradigms.} 
We report VTAB-1k~\cite{zhai2019large} accuracy using ViT-Base~\cite{dosovitskiy2021image} as the frozen backbone, pretrained with MAE~\cite{he2022masked} and MoCo v3~\cite{chen2021empirical} use the same settings.
}
\label{table:mae_moco}

\begin{adjustbox}{width=0.82\textwidth,center}
\begin{tabular}{c||r|rrr||r|rrr} 
\Xhline{4\arrayrulewidth}
\rowcolor{gray!20}
Pretrained & \multicolumn{4}{c||}{MAE~\cite{he2022masked}} & \multicolumn{4}{c}{MoCo v3~\cite{chen2021empirical}} \\
\rowcolor{gray!20}
Method & Tuned(\%) & \textit{Natural} & \textit{Specialized} & \textit{Structured} & Tuned(\%) & \textit{Natural} & \textit{Specialized} & \textit{Structured} \\
\hline \hline
Full \textcolor{lightgray}{\scriptsize{[CVPR22]}}\cite{iofinova2022well} & 100.00 & 59.31 & \textbf{79.68} & 53.82 & 100.00 & 71.95 & 84.72 & 51.98 \\
\hline
Linear \textcolor{lightgray}{\scriptsize{[CVPR22]}}\cite{iofinova2022well} & 0.04 & 18.87 & 53.72 & 23.70 & 0.04 & 67.46 & 81.08 & 30.33 \\
Partial-1 \textcolor{lightgray}{\scriptsize{[NeurIPS14]}}\cite{yosinski2014transferable} & 8.30 & 58.44 & 78.28 & 47.64 & 8.30 & 72.31 & 84.58 & 47.89 \\
\hline
Bias \textcolor{lightgray}{\scriptsize{[NeurIPS17]}}\cite{rebuffi2017learning} & 0.16 & 54.55 & 75.68 & 47.70 & 0.16 & 72.89 & 81.14 & 53.43 \\
Adapter \textcolor{lightgray}{\scriptsize{[NeurIPS20]}}\cite{cai2020tinytl} & 0.87 & 54.90 & 75.19 & 38.98 & 1.12 & 74.19 & 82.66 & 47.69 \\
\hline
VPT-S \textcolor{lightgray}{\scriptsize{[ECCV22]}}\cite{jia2022vpt} & 0.05 & 39.96 & 69.65 & 27.50 & 0.06 & 67.34 & 82.26 & 37.55 \\
VPT-D \textcolor{lightgray}{\scriptsize{[ECCV22]}}\cite{jia2022vpt} & 0.31 & 36.02 & 60.61 & 26.57 & 0.22 & 70.27 & 83.04 & 42.38 \\
GPT \textcolor{lightgray}{\scriptsize{[arXiv24]}}\cite{yoo2023improving} & 0.05 & 47.61 & 76.86 & 36.80 & 0.06 & 74.84 & 83.38 & 49.10 \\
VFPT \textcolor{lightgray}{\scriptsize{[NeurIPS24]}}\cite{zeng2024visual} & 0.38 & 53.59 & 77.75 & 36.15 & 0.22 & 77.47 & 85.76 & 58.74 \\
LoR-VP \textcolor{lightgray}{\scriptsize{[ICLR25]}}\cite{jin2025lor} & 0.40 & 51.23 & 74.21 & 33.16 & 0.29 & 72.28 & 82.89 & 46.92 \\
DA-VPT \textcolor{lightgray}{\scriptsize{[CVPR25]}}\cite{ren2025da-vpt} & — & \underline{62.14} & 79.14 & \underline{54.31} & — & 74.24 & 83.21 & 55.23 \\
ViaPT \textcolor{lightgray}{\scriptsize{[ACMMM25]}}\cite{xiao2025visual} & 0.36 & 54.26 & 78.01 & 37.52 & 0.30 & \underline{79.12} & \underline{86.81} & \underline{60.05} \\
\rowcolor{cvprblue!15}
\textbf{Ours} & 0.33 & \textbf{64.01} & \underline{79.20} & \textbf{55.29} & 0.29 & \textbf{79.91} & \textbf{87.15} & \textbf{61.06} \\
\hline
\end{tabular}
\end{adjustbox}
\end{table*}

\begin{table}[!ht]
  \centering
  \caption{\textbf{VTAB-1k results using Swin-Base pretrained on ImageNet-21k.}}
  \label{tab:performance_vtab1k}
  \begin{adjustbox}{width=0.94\linewidth,center}
  \begin{tabular}{c||c|ccc}
    \Xhline{1pt}
    \rowcolor{gray!20}
    Swin-Base~\cite{liu2021swin} & Tuned (\%) & \textit{Natural} & \textit{Specialized} & \textit{Structured} \\
    \hline \hline
    Full~\cite{iofinova2022well} & 100.00 & 79.10 & 86.21 & 59.65 \\
    Linear~\cite{iofinova2022well} & 0.06 & 73.52 & 80.77 & 33.52 \\
    Bias~\cite{rebuffi2017learning} & 0.29 & 74.19 & 80.14 & 42.42 \\
    VPT-D~\cite{jia2022vpt} & 0.25 & 76.78 & 83.33 & 51.85 \\
    E$^2$VPT~\cite{han2023e2vpt} & 0.21 & 83.31 & 84.95 & 57.35 \\
    SA$^2$VP~\cite{pei2024sa2vp} & 0.29 & 80.81 & \underline{86.30} & 60.03 \\
    VFPT~\cite{zeng2024visual} & 0.27 & 84.53 & 86.15 & 58.21 \\
    LoR-VP~\cite{jin2025lor} & 0.29 & 83.51 & 85.22 & 57.61 \\
    ViaPT~\cite{xiao2025visual} & 0.27 & \underline{85.29} & 86.21 & \textbf{62.24} \\
    \rowcolor{cvprblue!15}
    \textbf{Ours} & 0.26 & \textbf{85.40} & \textbf{87.11} & \underline{61.97} \\
    \Xhline{1pt}
  \end{tabular}
  \end{adjustbox}
\end{table}

\subsection{Experimental Setup}

We evaluate PIB in the standard frozen-backbone adaptation setting on VTAB-1k, following the three-group split of Natural, Specialized, and Structured. For all experiments, the pretrained backbone is kept frozen, and only the task head together with the prompt-related parameters are optimized. We report average classification accuracy over each VTAB-1k group. To test generality, we consider supervised ViT-Base/16, MAE- and MoCo v3-pretrained ViT-Base, and ImageNet-21k-pretrained Swin-Base. Unless otherwise stated, all compared methods follow the same evaluation protocol. Optimizer settings, schedules, batch size, training epochs, hardware, and other implementation details are provided in Appendix~\ref{app:implementation}.

\subsection{Main Results}

Tables~\ref{tab:vit}--\ref{tab:performance_vtab1k} summarize the comparisons across architectures and pretraining paradigms. Overall, PIB achieves consistently strong performance with a low tuning budget, showing that explicitly regulating layer-wise information allocation is effective for frozen-backbone adaptation.

\paragraph{Results on ViT-Base/16.}
As shown in Table~\ref{tab:vit}, PIB achieves the strongest overall performance on the standard ViT-Base/16 setting. The gains are especially clear on VTAB-1k, where our method improves all three splits and is particularly effective on the Structured subset. This suggests that the advantage of PIB does not come merely from stronger prompt parameterization, but from learning a more principled cross-layer information path.

\paragraph{Results across different pretraining paradigms.}
Table~\ref{table:mae_moco} shows that the benefit of PIB is robust across both reconstruction-based and contrastive pretraining. Under MAE and MoCo v3, our method remains consistently competitive and achieves the best or near-best results across most settings. This indicates that the proposed framework is not tied to a specific upstream training recipe, but generalizes across different pretrained representation spaces.

\paragraph{Results on Swin-Base.}
Table~\ref{tab:performance_vtab1k} shows that PIB transfers beyond standard ViTs: on Swin-Base, it reaches 85.40, 87.11, and 61.97 on Natural, Specialized, and Structured with only 0.26\% tunable parameters. This verifies compatibility with hierarchical transformer architectures.

\subsection{Ablation Studies}

We next study which components of PIB are responsible for the observed gains. Table~\ref{table:ablation_main} reports the component ablation, while Table~\ref{table:ablation_design} isolates path regularization and layer-aware weighting.

\begin{table*}[!ht]
\caption{\textbf{Component ablation of PIB.}
We evaluate the contribution of each component in the proposed framework on VTAB-1k using the frozen ViT-Base backbone. 
Starting from  VPT, we progressively add the compression regularizer, the sufficiency regularizer, the cross-layer path regularizer, and the routing mechanism. 
Higher is better for all metrics.
}
\label{table:ablation_main}
\tiny
\begin{adjustbox}{width=0.88\textwidth,center}
\begin{tabular}{c||cccc||ccc} 
\Xhline{4\arrayrulewidth}
\rowcolor{gray!20}
Method & Compression & Sufficiency & Path & Routing & \textit{Natural} & \textit{Specialized} & \textit{Structured} \\
\hline \hline
 VPT &  &  &  &  & 81.6 & 84.7 & 59.8 \\
+ Compression only & \checkmark &  &  &  & 82.0 & 84.9 & 61.2 \\
+ Sufficiency only &  & \checkmark &  &  & 82.4 & 85.2 & 60.8 \\
+ Comp. + Suff. & \checkmark & \checkmark &  &  & 82.8 & 85.5 & 62.3 \\
+ Comp. + Suff. + Path & \checkmark & \checkmark & \checkmark &  & 83.0 & 85.8 & 63.1 \\
\rowcolor{cvprblue!15}
\textbf{Full PIB} & \checkmark & \checkmark & \checkmark & \checkmark & \textbf{83.2} & \textbf{86.0} & \textbf{63.6} \\
\hline
\end{tabular}
\end{adjustbox}
\vspace{-1.2em}
\end{table*}

\begin{table}[!ht]
\scriptsize
\caption{\textbf{Design ablation of PIB.} We isolate path regularization and layer-aware weighting on VTAB-1k.}
\label{table:ablation_design}
\begin{adjustbox}{width=\linewidth,center}
\begin{tabular}{c||cc||ccc}
\Xhline{4\arrayrulewidth}
\rowcolor{gray!20}
Variant & Path & Layer-aware & \textit{Natural} & \textit{Specialized} & \textit{Structured} \\
\hline \hline
Uniform weighting &  &  & 82.6 & 85.3 & 61.9 \\
Uniform + Path & \checkmark &  & 82.9 & 85.7 & 62.8 \\
Layer-aware only &  & \checkmark & 82.8 & 85.6 & 62.4 \\
\rowcolor{cvprblue!15}
\textbf{Layer-aware + Path} & \checkmark & \checkmark & \textbf{83.2} & \textbf{86.0} & \textbf{63.6} \\
\hline
\end{tabular}
\end{adjustbox}
\end{table}

\begin{figure*}[t]
    \centering
    \includegraphics[width=\textwidth]{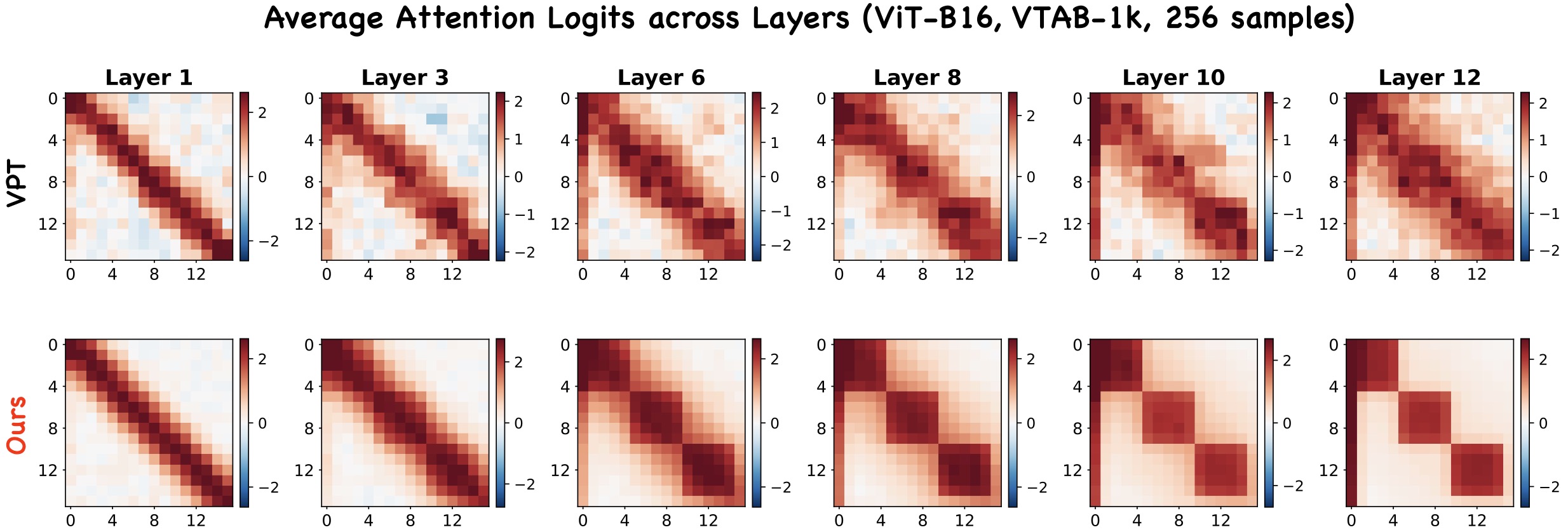}
    \caption{\textbf{Average attention logits across transformer depth.} We compare the layer-wise attention patterns of VPT and PIB on ViT-Base/16 over 256 VTAB-1k samples. VPT exhibits more fragmented and less stable attention evolution across depth, especially in middle and late layers. In contrast, PIB yields a smoother and more structured attention pattern, indicating a more coherent layer-wise refinement process.}
    \label{fig:attention_logits}
\end{figure*}

Table~\ref{table:ablation_main} shows that compression and sufficiency are complementary. Compression alone provides its largest gain on Structured tasks, where suppressing nuisance-heavy information is particularly important, while sufficiency yields broader gains by preserving discriminative semantics. Combining them improves all three groups. Adding the path regularizer further raises Structured accuracy from 62.3 to 63.1, supporting the need for coherent cross-layer refinement, and routing provides a smaller but consistent gain.

Table~\ref{table:ablation_design} isolates the two cross-layer design choices. Path regularization improves uniform weighting, and layer-aware weighting is also beneficial on its own. Their combination performs best across all three groups, indicating that a coherent global path and layer-dependent allocation play complementary roles. Additional component analysis is provided in Appendix~\ref{app:ablation}.

\subsection{Why Does Prompt Scaling Behave Non-Monotonically?}

Increasing prompt depth or length expands adaptation capacity but does not reliably improve transfer. We analyze both axes because depth controls how many transformer blocks receive prompts, whereas length controls the number of prompt tokens introduced at each prompted layer.

\begin{figure}[t]
    \centering
    \includegraphics[width=\linewidth]{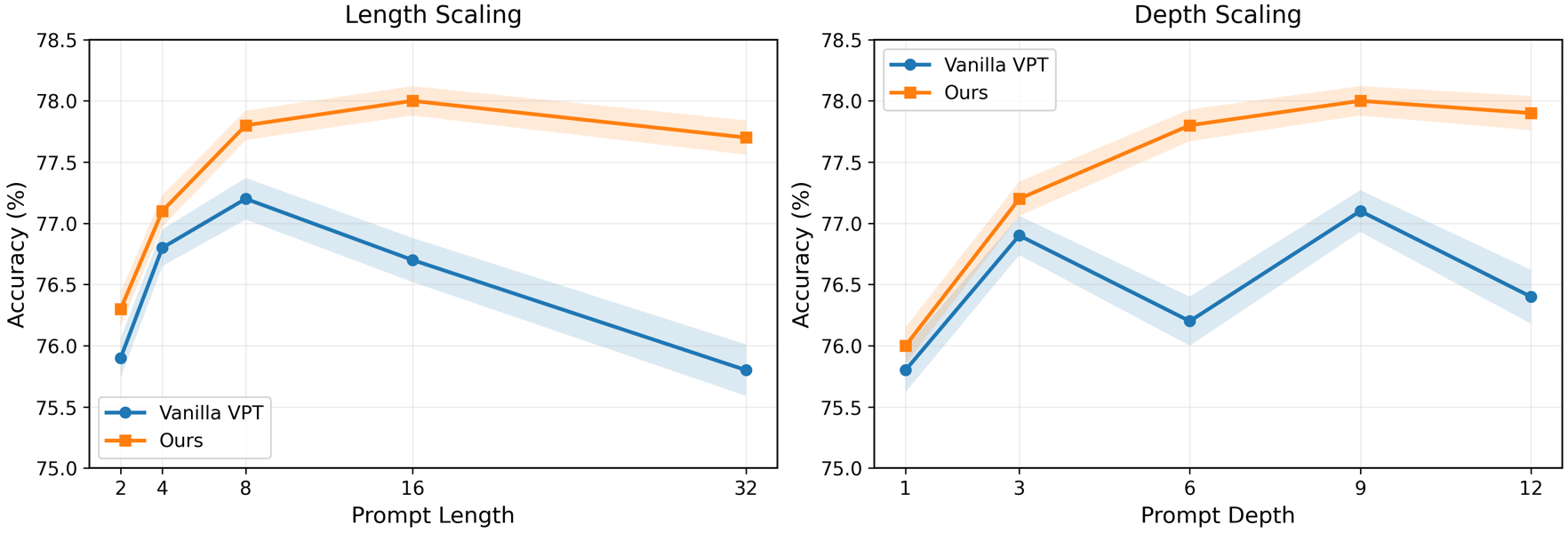}
    \caption{Depth and length scaling behavior of prompt-based adaptation. Increasing prompt capacity does not necessarily yield monotonic transfer gains. Standard VPT-style prompting exhibits noticeable instability as prompt depth or prompt length increases, whereas our method shows a smoother and more favorable scaling trend.}
    \label{fig:scaling_curves}
\end{figure}

\paragraph{Depth scaling.}
Prompting more layers is not uniformly beneficial because transformer depth carries different representational roles. Early blocks retain local visual evidence, while later blocks increasingly organize semantic structure. Injecting unconstrained prompts throughout the network can repeatedly preserve nuisance factors or disrupt this progression, producing the non-monotonic VPT curve in Figure~\ref{fig:scaling_curves}. PIB instead regulates the compression--sufficiency balance along depth and consequently exhibits a smoother trend.

\paragraph{Length scaling.}
Longer prompts similarly increase capacity without guaranteeing task-relevant information. Additional tokens may become redundant or amplify dataset-specific correlations when their information content is unconstrained. PIB remains more stable as length grows because its objective penalizes redundant intermediate structure while retaining class-discriminative evidence. Together, these results show that prompt scaling is governed by information allocation rather than capacity alone. Complete settings and additional discussion are provided in Appendix~\ref{app:ablation_scaling}.

\subsection{Attention Evolution Analysis}

Figure~\ref{fig:attention_logits} provides an independent view of how prompt-conditioned interactions evolve across depth. VPT displays a comparatively fragmented transition in middle and late layers, whereas PIB develops a smoother banded structure and more coherent deep-layer organization. The contrast is most visible after the intermediate blocks, where unconstrained prompts can propagate diffuse interactions. PIB does not simply increase attention magnitude; it reorganizes token relationships progressively, consistent with the path regularizer's role in discouraging abrupt deterioration of the compression--sufficiency trade-off. Attention entropy and prompt-to-patch analyses in Appendix~\ref{app:attention_vis} show the same progression from diverse early evidence to more selective late-stage structure.

\subsection{Layer-wise Information Allocation}

Beyond the main benchmark results, we further analyze how prompt-conditioned representations evolve across depth. To this end, we consider four complementary diagnostics: information trajectory, redundancy, class separability, and path score. Across all four metrics,  VPT exhibits less stable and less structured layer-wise behavior, while PIB consistently yields smoother refinement, lower redundancy, stronger separability, and a more ordered cross-layer path. These observations provide direct support for our central claim: the weakness of  VPT is not only reflected in final performance, but also in disordered layer-wise information evolution. In contrast, PIB improves prompt-based adaptation by making this cross-layer process less oscillatory, less redundant, and more discriminative. A representative visualization is shown in Figure~\ref{fig:intro_diag}. Appendix~\ref{app:attention_vis} provides attention entropy, average-logit, and prompt-to-patch analyses, including Figure~\ref{fig:attn_logits}; further layer-wise diagnostics in Appendix~\ref{app:diagnostics}.

\subsection{Qualitative Visualization}

\begin{figure*}[t]
    \centering
    \includegraphics[width=\textwidth]{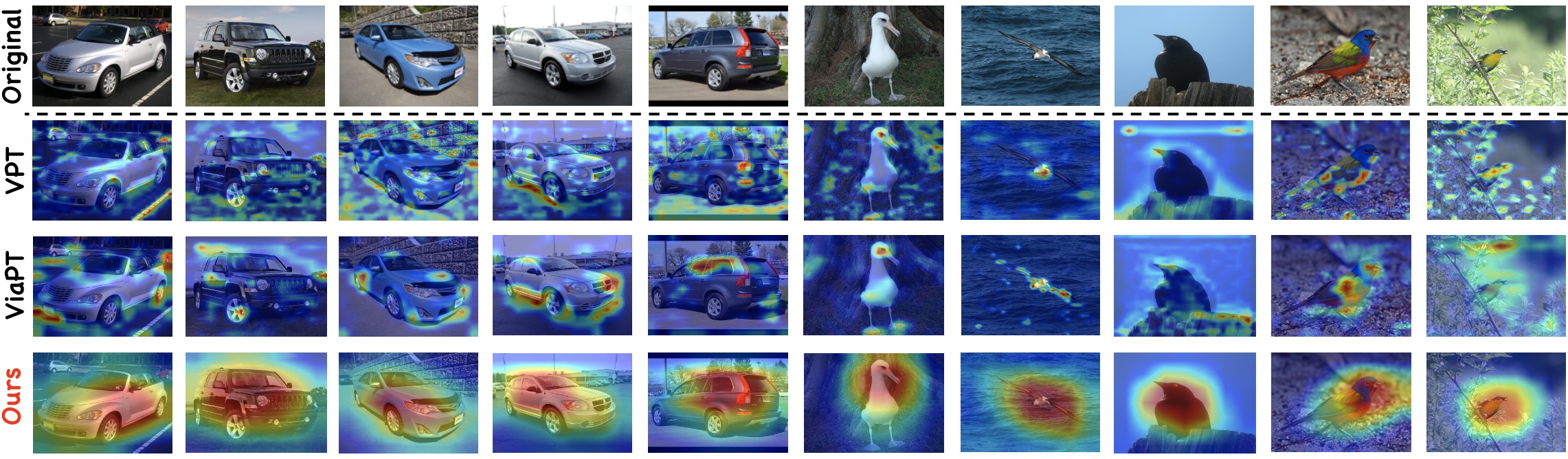}
    \caption{\textbf{Qualitative visualization of prompt attention. }
    Compared with VPT and ViaPT, our method produces more concentrated and semantically aligned responses on the foreground object, while reducing scattered activation on background regions. This pattern is consistent across both vehicle and bird categories, suggesting that PIB encourages prompt-conditioned representations to preserve task-relevant evidence and suppress redundant or nuisance-heavy information.}
    \label{fig:vis_analysis}
\end{figure*}

Figure~\ref{fig:vis_analysis} provides a qualitative comparison of prompt attention maps. VPT often responds to a mixture of foreground content and surrounding context, while PIB produces more object-centric and semantically aligned activations. This difference is especially clear in the bird examples, where background clutter and vegetation are strong sources of distraction: compared with both VPT~\cite{jia2022vpt} and ViaPT~\cite{xiao2025visual}, PIB places more emphasis on the discriminative object region and suppresses scattered responses on irrelevant surroundings. These patterns are consistent with our central hypothesis that prompt tuning benefits from explicitly regulating layer-wise information allocation. By progressively filtering nuisance-heavy signals while preserving task-relevant evidence, PIB leads to prompt-conditioned representations that are not only more interpretable, but also better aligned with robust semantic recognition.

\subsection{Does PIB Reduce Shortcut Reliance and Improve Robustness?}

A central claim of this paper is that the benefit of PIB is not limited to improving standard transfer accuracy. If the proposed framework indeed regulates layer-wise information allocation more effectively, then its advantage should also appear in settings where shortcut-heavy adaptation becomes unreliable, such as foreground-background imbalance, corrupted inputs, distribution shift, and low-data transfer. We therefore evaluate PIB from two complementary perspectives: shortcut reliance and robustness under challenging transfer conditions. We first test whether PIB reduces dependence on shortcut-prone visual cues. As shown in Table~\ref{table:shortcut_robustness}, PIB achieves a higher accuracy than VPT under the standard setting (78.1 vs.\ 77.4), and the gain becomes larger when evaluation emphasizes foreground-dominant information (75.6 vs.\ 73.1). In contrast, PIB performs worse under the background-only setting (63.4 vs.\ 67.8). We interpret this result together with the standard and foreground-only gains: under the same evaluation construction, PIB retains stronger task-relevant foreground evidence while deriving less predictive benefit from background-only inputs. This supports, rather than alone proves, reduced shortcut reliance.
\begin{table}[!ht]
\scriptsize
\caption{\textbf{Shortcut reliance and robustness evaluation.}
Higher is better for all metrics, while Drop$\downarrow$ denotes the average degradation from the standard setting.
}
\label{table:shortcut_robustness}

\begin{adjustbox}{width=\linewidth,center}
\begin{tabular}{c||ccc||cccc} 
\Xhline{4\arrayrulewidth}
\rowcolor{gray!20}
Method & \multicolumn{3}{c||}{Shortcut Reliance} & \multicolumn{4}{c}{Robustness} \\
\rowcolor{gray!20}
 & Std. & FG-only & BG-only & Corr. & Shift & Few-shot & Drop$\downarrow$ \\
\hline \hline
 VPT & 77.4 & 73.1 & 67.8 & 71.6 & 69.8 & 66.9 & 8.0 \\
\rowcolor{cvprblue!15}
\textbf{Ours} & 78.1 & 75.6 & 63.4 & 74.8 & 73.2 & 70.5 & 5.3 \\
\hline
\end{tabular}
\end{adjustbox}
\vspace{-1.2em}
\end{table}
We further evaluate robustness under corruption, domain shift, and few-shot transfer. PIB consistently outperforms VPT across all three settings, improving accuracy from 71.6 to 74.8 under corruption, from 69.8 to 73.2 under domain shift, and from 66.9 to 70.5 under few-shot adaptation. More importantly, PIB exhibits a substantially smaller average degradation from the standard setting, reducing the drop from 8.0 points to 5.3 points. This result indicates that the improvement brought by PIB is not confined to the in-domain regime; instead, it yields a more stable transfer profile once the evaluation condition becomes more challenging. By explicitly constraining what prompts preserve and discard across layers, PIB progressively suppresses redundant information while retaining more task-relevant evidence. The learned prompt-conditioned representations are not only more discriminative, but also less fragile under perturbation and distribution shift. This is consistent with our main thesis that prompt-based adaptation benefits from regulating the layer-wise compression-sufficiency trade-off.

\section{Conclusion}

In this paper, we study prompt-based vision foundation model adaptation from an information-allocation perspective and argue that its limitation is not simply insufficient prompt capacity, but the lack of principled control over what prompt-conditioned representations preserve and suppress across layers. Based on this view, we present Prompted Information Bottlenecks (PIB), which regularize prompt adaptation through complementary compression and sufficiency objectives together with a cross-layer minimal-sufficient path. Extensive experiments show that PIB improves parameter-efficient transfer across different backbones and pretraining paradigms, while also yielding more stable layer-wise behavior and stronger robustness. We hope this work provides a useful step toward a more principled understanding of frozen VFM adaptation. More broadly, these findings suggest that parameter-efficient adaptation should be understood not only in terms of how few parameters are updated, but also in terms of how adaptation reshapes intermediate representations.


\clearpage

\bibliographystyle{ACM-Reference-Format}
\bibliography{ref}
\clearpage
\appendix

\renewcommand{\thefigure}{S\arabic{figure}}
\setcounter{figure}{0}
\renewcommand{\thetable}{S\arabic{table}}
\setcounter{table}{0}

\makeatletter
\@ifundefined{theHfigure}{}{\renewcommand{\theHfigure}{S\arabic{figure}}}
\@ifundefined{theHtable}{}{\renewcommand{\theHtable}{S\arabic{table}}}
\makeatother

\clearpage
\twocolumn[{%
\centering
\vspace*{-0.8em}
{\LARGE\bfseries Appendix\par}
\vspace{0.35em}
{\large\bfseries Additional Implementation Details, Analyses, and Results\par}
\vspace{0.55em}
\noindent\rule{\textwidth}{1.1pt}
\vspace{0.8em}
}]


\noindent
This appendix provides additional implementation details, extended empirical results, and further discussion of our method, organized as follows:
\vspace{0.2em}

\begin{itemize}[leftmargin=1.5em,itemsep=0.35em,topsep=0.3em]
    \item \textbf{Section~A} provides additional implementation details, including backbone settings, prompt injection, optimization protocol, and the implementation of the proposed compression, sufficiency, path, and routing components.

    \item \textbf{Section~B} presents detailed per-task benchmark results on FGVC, HTA, and VTAB-1k.

    \item \textbf{Section~C} summarizes dataset details and evaluation protocols for all benchmarks used in this work.

    \item \textbf{Section~D} reports extended ablation studies, including component-wise analysis, design ablations, and scaling behavior with respect to prompt depth and prompt length.

    \item \textbf{Section~E} provides more detailed diagnostic results for layer-wise information allocation across different backbones, task groups, and pretraining paradigms.

    \item \textbf{Section~F} presents additional attention visualizations and interpretability analyses, including attention entropy, average attention logits, and prompt-to-patch attention patterns.

    \item \textbf{Section~G} discusses the theoretical role and limitations of the practical surrogates used in PIB.

    \item \textbf{Section~H} analyzes efficiency, complexity, and overhead, with particular emphasis on parameter efficiency and training/inference cost.

    \item \textbf{Section~I} gives an open discussion of how layer-wise information allocation may extend to heterogeneous supervision, structured temporal representations, and generative or knowledge-intensive multimodal tasks.

    \item \textbf{Section~J} discusses failure cases, current limitations, and possible directions for future work.
\end{itemize}

\vspace{0.5em}

\section{Additional Implementation Details}
\label{app:implementation}

This section provides additional implementation details for reproducing the proposed Prompted Information Bottleneck (PIB) framework. Unless otherwise stated, all experiments follow the same frozen-backbone adaptation protocol as described in the main paper: the pretrained visual backbone is kept fixed, while only the task head, prompt-related parameters, and, when enabled, the routing variables are optimized.

\subsection{Backbone Settings and Prompt Injection}
\label{app:backbone_prompt}

We evaluate PIB on three backbone families in order to test its generality across different representation spaces and transformer designs: (i) supervised ViT-Base/16, (ii) ViT-Base pretrained with MAE and MoCo v3, and (iii) Swin-Base pretrained on ImageNet-21k. For all ViT-based models, prompts are injected in a deep prompting manner, i.e., a learnable prompt set is associated with each transformer block. Let the backbone have $L$ blocks and hidden dimension $d$. The prompt parameters are denoted by $\{P_l\}_{l=1}^L$, where $P_l \in \mathbb{R}^{M_l \times d}$ is the learnable prompt matrix at layer $l$.

For ViT-style backbones, the prompts are concatenated to the token sequence before entering each frozen transformer block, following the standard prompt tuning protocol. The class token is preserved throughout the network, and prompt tokens are removed after each block so that only image tokens and the class token are propagated to the next layer together with the newly injected prompts. For Swin-Base, which follows a hierarchical design, we apply the same principle stage-wise: prompt tokens are attached to the token sequence within each stage while preserving the original window-based attention computation of the frozen backbone. In all cases, the backbone weights, including patch embedding, self-attention, MLP layers, normalization layers, and positional embeddings, remain frozen during downstream adaptation.

Unless otherwise noted, prompt parameters are initialized from a zero-mean Gaussian distribution with small standard deviation. We found this more stable than large random initialization, especially when the path regularizer is enabled. The task head is implemented as a single linear classifier attached to the final backbone representation. For ViT-based models, the classifier operates on the final class token. For Swin-Base, the classifier is applied to the pooled output representation used by the original backbone.

\subsection{Optimization Protocol}
\label{app:optimization}

We optimize PIB with AdamW for all experiments. The backbone is frozen throughout training, and gradients are computed only for the prompt parameters, the classification head, and the routing variables when the routed PIB variant is used. The default training objective is
\begin{equation}
\mathcal{L}_{\text{total}}
=
\mathcal{L}_{\text{task}}
+
\mathcal{L}_{\text{PIB}}
+
\eta \mathcal{L}_{\text{path}},
\end{equation}
or
\begin{equation}
\mathcal{L}_{\text{total}}
=
\mathcal{L}_{\text{task}}
+
\mathcal{L}_{\text{PIB}}^{\text{route}}
+
\eta \mathcal{L}_{\text{path}},
\end{equation}
when layer-aware routing is enabled.

For the task loss, we use standard cross-entropy. The optimizer hyperparameters are selected on the validation split using a small search budget shared across compared prompt-based methods. In practice, the learning rate is chosen from a compact grid, and we found that prompt-based adaptation is generally stable within the standard PEFT range. We use a cosine decay schedule with linear warm-up at the beginning of training. Warm-up is particularly helpful for PIB because the compression and sufficiency terms may otherwise dominate too early before the classifier head becomes reasonably calibrated.

To keep the comparison fair across benchmarks, we use the same optimization family across all datasets and vary only a small number of coarse hyperparameters, namely the learning rate, total training epochs, and batch size. Weight decay is applied to the trainable parameters, while no gradient is propagated into the frozen backbone. We do not use label smoothing, model EMA, or test-time augmentation unless explicitly stated in the corresponding experiment. Gradient clipping is applied when training becomes unstable under large prompt length or strong regularization.

All reported main results are obtained from the best validation checkpoint under the same model selection rule across methods. Unless otherwise stated, each configuration is run with multiple random seeds, and the final performance is reported as the average test accuracy under the selected hyperparameters.

\subsection{Implementation of the Compression Loss}
\label{app:compression_impl}

The compression term is designed to reduce redundant or nuisance-heavy propagation in layer-wise hidden representations. For each prompted layer $l$, let $H_l \in \mathbb{R}^{N_l \times d}$ denote the hidden representation after prompt-conditioned processing, where $N_l$ is the number of tokens and $d$ is the embedding dimension. We first apply row-wise $\ell_2$ normalization:
\begin{equation}
\tilde{H}_l(i,:) = \frac{H_l(i,:)}{\|H_l(i,:)\|_2 + \epsilon_{\text{norm}}},
\end{equation}
where $\epsilon_{\text{norm}}$ is a small numerical constant.

We then compute the empirical token correlation matrix
\begin{equation}
C_l = \frac{1}{N_l} \tilde{H}_l^\top \tilde{H}_l.
\end{equation}
The compression loss is implemented as the squared Frobenius norm of the off-diagonal entries:
\begin{equation}
\mathcal{L}^{(l)}_{\text{comp}}
=
\| C_l - I_d \|^2_{F,\text{off}}.
\end{equation}
In implementation, we explicitly mask out the diagonal entries before summation:
\begin{equation}
\mathcal{L}^{(l)}_{\text{comp}}
=
\sum_{i \neq j}
\left(C_l(i,j)\right)^2.
\end{equation}
This avoids penalizing the self-correlation terms and ensures that the objective focuses on redundancy among representation dimensions rather than shrinking the representation norm itself.

We apply this loss to the prompt-conditioned hidden states at every prompted layer. In practice, the magnitude of $\mathcal{L}^{(l)}_{\text{comp}}$ varies across depth because early layers carry more local variation, while deeper layers are more semantically organized. For this reason, the compression coefficients $\lambda_l$ are allowed to vary across layers. In the basic PIB variant, $\lambda_l$ follows a fixed schedule; in the routed variant, it is further modulated by the learned routing gate.

\subsection{Implementation of the Sufficiency Loss}
\label{app:sufficiency_impl}

Compression alone may remove discriminative information that is necessary for the downstream task. To prevent over-compression, PIB introduces a sufficiency term based on mini-batch class separability. For each sample $i$ at layer $l$, we obtain a pooled representation $h_{l,i} \in \mathbb{R}^{d}$ from the prompt-conditioned hidden state. For ViT-style backbones, we use the class token representation by default. For hierarchical backbones such as Swin, we use the pooled stage output that is fed to the classifier head.

Let $\mathcal{B}_c$ denote the set of samples of class $c$ in a mini-batch, and let $\mu_{l,c}$ denote the class mean at layer $l$. The within-class and between-class scatter are computed as
\begin{equation}
S^{(l)}_{\text{intra}}
=
\sum_c \sum_{i \in \mathcal{B}_c}
\|h_{l,i} - \mu_{l,c}\|_2^2,
\end{equation}
and
\begin{equation}
S^{(l)}_{\text{inter}}
=
\sum_{c \neq c'}
\|\mu_{l,c} - \mu_{l,c'}\|_2^2.
\end{equation}
The layer-wise sufficiency loss is
\begin{equation}
\mathcal{L}^{(l)}_{\text{suff}}
=
\frac{S^{(l)}_{\text{intra}}}{S^{(l)}_{\text{inter}} + \epsilon_{\text{suff}}},
\end{equation}
where $\epsilon_{\text{suff}}$ is a small constant for numerical stability.

In practice, the sufficiency term is computed only over classes that appear at least twice in a mini-batch, since a single example cannot define a meaningful within-class scatter. When a class appears only once, it is excluded from the intra-class term of that batch. We found this implementation detail important for low-shot settings and small-batch runs. To reduce variance, the sufficiency loss is averaged across valid classes within the batch before being combined across layers.

\subsection{Cross-Layer Path Regularization}
\label{app:path_impl}

The path regularizer enforces the idea that prompt-conditioned representations should evolve along a coherent minimal-sufficient path rather than oscillate arbitrarily across depth. For each layer, we define a quality score
\begin{equation}
q_l = \mathcal{L}^{(l)}_{\text{comp}} + \alpha_l \mathcal{L}^{(l)}_{\text{suff}},
\end{equation}
where $\alpha_l$ balances the relative contribution of compression and sufficiency at layer $l$. The path regularizer is then written as
\begin{equation}
\mathcal{L}_{\text{path}}
=
\sum_{l=1}^{L-1}
\max\left(0, q_{l+1} - q_l - \delta_l\right),
\end{equation}
where $\delta_l \ge 0$ is a tolerance margin.

In implementation, we use a small non-negative margin to avoid over-constraining adjacent layers. This is important because neighboring layers in a frozen transformer do not serve identical representational roles. The path term is thus not intended to enforce strict monotonicity, but rather to prevent abrupt deterioration of the compression--sufficiency trade-off. Empirically, this formulation yields smoother layer-wise dynamics than either unconstrained training or an overly strict monotonic penalty.

For efficiency, all layer-wise quality scores are computed in the same forward pass used for task prediction. No auxiliary branch or teacher network is introduced. The path loss is therefore inexpensive relative to full fine-tuning and adds only a small overhead on top of standard deep prompt tuning.

\subsection{Layer-Aware Routing}
\label{app:routing_impl}

The routed PIB variant introduces a learnable gate $r_l \in [0,1]$ for each prompted layer:
\begin{equation}
\mathcal{L}_{\text{PIB}}^{\text{route}}
=
\sum_{l=1}^{L}
\left(
r_l \lambda_l \mathcal{L}^{(l)}_{\text{comp}}
+
(1-r_l)\gamma_l \mathcal{L}^{(l)}_{\text{suff}}
\right).
\end{equation}
We parameterize $r_l$ through an unconstrained scalar $a_l$ and apply a sigmoid transformation,
\begin{equation}
r_l = \sigma(a_l),
\end{equation}
so that the routing weight remains in $[0,1]$ throughout training.

The routing variables are optimized jointly with the prompts and the classifier head. To avoid degenerate behavior in which all layers collapse to the same routing pattern, we initialize the routing logits near the center of the sigmoid function, which allows the model to adapt toward either preservation-dominant or compression-dominant behavior depending on the downstream task. In our experiments, this routed version serves as a refinement rather than a replacement of the base PIB formulation. It typically provides smaller but consistent gains on top of the compression, sufficiency, and path objectives.

\subsection{Layer-Wise Coefficient Design}
\label{app:coefficients}

PIB uses four groups of layer-wise coefficients: $\lambda_l$, $\gamma_l$, $\alpha_l$, and $\delta_l$. We keep these schedules lightweight and low-dimensional so that the method does not rely on expensive per-task tuning.

For the basic setting, $\lambda_l$ and $\gamma_l$ follow a simple depth-aware schedule: early layers place relatively more emphasis on sufficiency to preserve local evidence, while deeper layers place relatively more emphasis on compression to suppress nuisance-heavy or redundant information. This design matches the motivation in the main paper that early layers should remain detail-preserving and later layers should become increasingly selective. Similarly, $\alpha_l$ is set so that the path score remains numerically balanced across layers, and $\delta_l$ is chosen as a small tolerance term rather than a strong monotonic constraint.

To further reduce the sensitivity of these coefficients, we first normalize the layer-wise loss magnitudes within each mini-batch before aggregation. This stabilizes training across datasets with different class cardinality and representation statistics, and makes the resulting schedules transferable across backbone families. In the routed variant, the learned gates provide an additional adaptive mechanism on top of these fixed schedules.

\subsection{Data Preprocessing and Augmentation}
\label{app:data_preprocessing}

All images are resized to the input resolution expected by the corresponding pretrained backbone. Unless required otherwise by the dataset protocol, we use standard ImageNet-style normalization. During training, we apply a lightweight augmentation pipeline consisting of random resized crop and horizontal flip. We intentionally avoid heavy augmentation because the goal of this work is to study prompt-induced information allocation under a controlled frozen-backbone setting rather than maximize benchmark performance through aggressive data transformation.

At evaluation time, we use a single center crop or the standard test preprocessing protocol defined by the benchmark. No multi-crop evaluation or test-time ensembling is used in the main results. This keeps the comparison consistent with prior prompt-tuning baselines and ensures that gains are attributable to the proposed information regularization rather than inference-time heuristics.

\subsection{Batch Construction and Numerical Stability}
\label{app:batch_stability}

Several implementation details are important for stable optimization of PIB. First, because the sufficiency term depends on class statistics within a mini-batch, we use class-balanced batching whenever the benchmark protocol allows it. When this is not feasible, we rely on standard random mini-batches and exclude singleton classes when computing the within-class term, as discussed earlier. Second, we use small numerical constants in the normalization and denominator terms of both the compression and sufficiency objectives to avoid instability in early training.

Third, we found it beneficial to delay the full influence of the regularization terms during the warm-up phase. Concretely, in the first few epochs, the classifier head is allowed to stabilize before the path regularizer reaches its full strength. This avoids poor local minima caused by over-regularizing noisy early representations. Finally, all losses are accumulated in full precision even when mixed-precision training is used for the forward and backward passes.

\subsection{Hardware and Software}
\label{app:hardware}

All experiments are implemented in PyTorch and trained on NVIDIA GPUs. We use mixed-precision training to reduce memory consumption and accelerate optimization. The exact hardware configuration depends on the benchmark scale, but the overall setting remains modest because only a very small fraction of parameters is trainable. This is consistent with the parameter-efficient nature of PIB and with the frozen-backbone protocol used throughout the paper.

The codebase is built on top of a standard vision transformer training framework, with PIB added as a lightweight regularization module that hooks into the hidden states of prompted layers. The additional memory overhead mainly comes from storing layer-wise hidden representations for computing the compression and sufficiency losses. In practice, this overhead is small compared with end-to-end fine-tuning of the full backbone.

\section{Detailed Main Results}
\label{app:detailed_results}

In this section, we provide a more fine-grained view of the benchmark results summarized in the main comparison tables. While the main paper reports benchmark-level averages for FGVC, HTA, and VTAB-1k, the detailed tables in this appendix reveal how the proposed Prompted Information Bottleneck (PIB) behaves on individual downstream tasks. Overall, the per-task results are consistent with the central conclusion of the main paper: explicitly regulating layer-wise information allocation yields stronger and more stable transfer performance under the frozen-backbone setting. In particular, PIB achieves the best overall benchmark-level performance on the standard ViT-B/16 setting, reaching 92.1 on FGVC, 93.01 on HTA, and 77.33 on VTAB-1k, while maintaining a low tuning budget.

\subsection{Per-task Results on FGVC}
\label{app:fgvc_results_text}

Table~\ref{tab:fgvc_per_task} reports the detailed results on the five FGVC datasets. Consistent with the benchmark-level summary in the main paper, PIB achieves the strongest average performance, with a mean accuracy of 92.10. More importantly, the gains are not concentrated on only one or two datasets. Instead, our method delivers consistent improvements across all five fine-grained recognition tasks, including CUB-200, NABirds, Oxford Flowers, Stanford Dogs, and Stanford Cars. This behavior is important because FGVC tasks are particularly sensitive to whether the adaptation method preserves subtle category-specific evidence while suppressing irrelevant variation. The fact that PIB improves all five datasets suggests that the proposed compression--sufficiency regularization does not merely increase prompt expressiveness, but improves how useful visual evidence is retained and refined across transformer depth. This observation is also aligned with the main paper’s broader claim that PIB leads to more robust fine-grained recognition behavior rather than only higher average accuracy.

Compared with prior prompt-based baselines such as VPT-Deep and E$^2$VPT, the advantage of PIB is especially notable on the more recognition-sensitive FGVC datasets. In particular, the improvement on Stanford Cars is relatively pronounced, while gains on CUB-200, NABirds, and Stanford Dogs further indicate that PIB is effective in settings where class boundaries are narrow and intra-class variation remains substantial. Together, these results support the interpretation that PIB better preserves discriminative local cues in earlier layers while avoiding the over-propagation of redundant or nuisance-heavy information into later layers. This is consistent with the information-allocation view introduced in the main paper.

\subsection{Per-task Results on HTA}
\label{app:hta_results_text}

Table~\ref{tab:hta_per_task} presents the detailed results on the ten HTA datasets. PIB again achieves the strongest overall mean, reaching 93.01, which matches the best benchmark-level result reported in the main paper. Unlike FGVC, HTA covers a more heterogeneous collection of visual domains, including texture recognition, natural object recognition, food classification, traffic signs, and digit recognition. The detailed results show that PIB remains competitive across this broader transfer spectrum and achieves the best score on every task reported in the table.

This finding is important for two reasons. First, it shows that the benefit of PIB is not restricted to fine-grained recognition alone. Even when the downstream tasks vary substantially in visual statistics and semantic structure, explicitly regulating what prompt-conditioned representations should preserve and suppress still leads to stronger transfer. Second, the improvement pattern suggests that PIB is able to adapt without over-specializing to one narrow task family. Gains are visible not only on fine-grained datasets such as CUB-200, NABirds, Dogs, and Flowers, but also on more heterogeneous tasks such as DTD, Food-101, CIFAR-100, CIFAR-10, GTSRB, and SVHN. This broad effectiveness is consistent with the main paper’s conclusion that PIB improves frozen-backbone adaptation through a more principled cross-layer information path rather than through a task-specific prompt engineering trick.

Among all HTA tasks, the relative improvements are particularly clear on DTD, CIFAR-100, GTSRB, and SVHN, where robust suppression of nuisance-heavy or shortcut-prone cues is especially beneficial. At the same time, PIB continues to deliver strong results on already high-performing tasks such as Flowers and CIFAR-10, indicating that the proposed regularization does not trade away in-domain accuracy in order to obtain robustness elsewhere. Instead, it improves the overall transfer profile in a balanced manner. This observation is in line with the main text, which emphasizes that PIB is effective not only in average accuracy but also in robustness-oriented settings.

\subsection{Per-task Results on VTAB-1k}
\label{app:vtab_results_text}

We further report the detailed VTAB-1k results in Table~\ref{tab:vtab_per_task}. This table makes the pattern already visible in the main paper much clearer. PIB achieves 82.98 on Natural, 85.83 on Specialized, and 63.18 on Structured, yielding the strongest overall VTAB-1k performance among the compared methods. Most importantly, the gain is consistently observed across all three groups and is largest on the Structured subset, which is exactly the trend highlighted in the main paper.

The Natural tasks show that PIB remains strong even when the target tasks stay relatively close to standard natural-image recognition. On this subset, PIB improves over strong prompt-based baselines while preserving a favorable parameter-efficiency profile. On the Specialized subset, PIB again achieves the highest mean, indicating that the proposed framework transfers well to medical, satellite, and remote-sensing-style recognition tasks where the visual distribution differs substantially from standard ImageNet-style pretraining. The most striking improvement appears on the Structured subset. Here, PIB yields the largest margin, which strongly supports our core hypothesis: when the task depends more heavily on compositional reasoning, geometry, or relational structure, prompt tuning benefits substantially from explicit control over layer-wise information allocation rather than simply increasing prompt capacity.

A closer look at the per-task VTAB results further reinforces this interpretation. On several structure-sensitive tasks, such as CLEVR, KITTI Distance, dSprites, and SmallNORB, PIB consistently improves over VPT-style baselines. These tasks are more vulnerable to unstable intermediate refinement and to the persistence of non-essential variation across layers. The stronger performance of PIB on these tasks therefore provides direct empirical support for the information-allocation argument made in the main paper: the weakness of vanilla VPT is not merely insufficient prompt capacity, but the lack of principled control over what intermediate prompt-conditioned representations should retain or discard as depth increases.

\subsection{Connection to the Main Results}
\label{app:main_results_connection}

Taken together, the detailed per-task tables support the same three conclusions stated in the main paper. First, PIB is effective across a broad range of downstream tasks rather than only on a small subset. Second, its advantage becomes more pronounced in harder transfer settings, especially on the Structured part of VTAB-1k. Third, these gains are achieved without sacrificing parameter efficiency, which remains one of the main strengths of the proposed framework. In other words, the detailed results confirm that the improvements reported in the main paper are not artifacts of averaging, but reflect a consistent and interpretable transfer advantage across individual datasets.

\begin{table*}[t]
  \centering
  \caption{\textbf{Per-task results on HTA (ViT-B/16, frozen).}}
  \label{tab:hta_per_task}
  \begin{adjustbox}{width=\textwidth,center}
  \begin{tabular}{l||ccccccccccc}
    \Xhline{1pt}
    \rowcolor{gray!20}
    \textbf{Methods} & \textbf{DTD} & \textbf{CUB-200} & \textbf{NABirds} & \textbf{Dogs} & \textbf{Flowers} & \textbf{Food-101} & \textbf{CIFAR-100} & \textbf{CIFAR-10} & \textbf{GTSRB} & \textbf{SVHN} & \textbf{Mean} \\
    \hline \hline
    Full FT                                 & 64.3 & 87.3 & 82.7 & 89.4 & 98.8 & 84.9 & 68.9 & 97.4 & \second{97.1} & 87.4 & 85.8 \\
    Head FT                                 & 63.2 & 85.3 & 75.9 & 86.2 & 97.9 & 84.4 & 63.4 & 96.3 & 68.0 & 36.6 & 75.7 \\
    Adapter~\cite{houlsby2019parameter}     & 62.7 & 87.1 & \second{84.3} & 89.8 & 98.5 & 86.0 & 74.2 & \third{97.7} & 91.1 & 36.3 & 80.8 \\
    VPT-Deep~\cite{jia2022vpt}              & 65.8 & \second{88.5} & \third{84.2} & \third{90.2} & \third{99.0} & 83.3 & 78.8 & 96.8 & 90.7 & 78.1 & 85.5 \\
    AdaptFormer~\cite{chen2022adaptformer}  & \second{74.4} & 84.7 & 75.2 & 84.7 & 97.9 & \second{89.1} & \second{91.4} & \second{98.8} & \third{97.0} & \second{96.5} & \second{89.0} \\
    DAM-VP~\cite{huang2023diversity}        & \third{73.1} & \third{87.5} & 82.1 & \second{92.3} & \second{99.2} & \third{86.9} & \third{86.9} & 90.6 & 87.9 & \third{88.1} & \third{88.5} \\
    \Xhline{1pt}
    \rowcolor{cvprblue!15}
    \textbf{Ours}                           & \best{79.2} & \best{91.0} & \best{87.3} & \best{93.0} & \best{99.6} & \best{91.1} & \best{93.8} & \best{99.2} & \best{98.4} & \best{97.5} & \best{93.01} \\
    \Xhline{1pt}
  \end{tabular}
  \end{adjustbox}
\end{table*}

\begin{table*}[t]
  \centering
  \caption{\textbf{Per-task results on FGVC (ViT-B/16, frozen).}}
  \label{tab:fgvc_per_task}
  \begin{adjustbox}{width=0.8\textwidth,center}
  \begin{tabular}{l||cccccc}
    \Xhline{1pt}
    \rowcolor{gray!20}
    \textbf{Methods} & \textbf{CUB-200} & \textbf{NABirds} & \textbf{Oxford Flowers} & \textbf{Stanford Dogs} & \textbf{Stanford Cars} & \textbf{Mean} \\
    \hline \hline
    Full FT                                 & 87.3 & 82.7 & 98.8 & 89.4 & \second{84.5} & 88.54 \\
    AdaptFormer~\cite{chen2022adaptformer} & 84.7 & 75.2 & 97.9 & 84.7 & 83.1 & 85.12 \\
    LoRA~\cite{hu2022lora}                  & 84.9 & 79.0 & 98.1 & 88.1 & 79.8 & 85.98 \\
    VPT-Shallow~\cite{jia2022vpt}           & 86.7 & 78.8 & 98.4 & \second{90.7} & 68.7 & 84.62 \\
    VPT-Deep~\cite{jia2022vpt}              & \third{88.5} & \third{84.2} & \third{99.0} & 90.2 & \third{83.6} & \third{89.11} \\
    E$^2$VPT~\cite{han2023e2vpt}            & \second{89.1} & \second{84.6} & \second{99.1} & \third{90.5} & 82.8 & \second{89.22} \\
    \Xhline{1pt}
    \rowcolor{cvprblue!15}
    \textbf{Ours}                           & \best{90.4} & \best{87.1} & \best{99.6} & \best{92.0} & \best{91.4} & \best{92.10} \\
    \Xhline{1pt}
  \end{tabular}
  \end{adjustbox}
\end{table*}

\begin{table*}[t]
  \centering
  \small
  \caption{\textbf{VTAB-1k per-task results (ViT-B/16, frozen).}
  Full: full finetune; Head: head finetune; AdaptF: AdaptFormer.}
  \label{tab:vtab_per_task}
  \begin{adjustbox}{width=0.8\textwidth,center}
  \begin{tabular}{l||cccccccc}
    \Xhline{1pt}
    \rowcolor{gray!20}
    \textbf{Datasets} & \textbf{Full} & \textbf{Head} & \textbf{AdaptF} & \textbf{LoRA} & \textbf{VPT-D} & \textbf{ExPRes} & \textbf{E$^{2}$VPT} & \textbf{Ours} \\
    \hline \hline
    \multicolumn{9}{l}{\textit{Natural}} \\
    CIFAR-100             & 68.9 & 63.4 & 70.8 & 67.1 & \second{78.8} & \third{78.0} & 78.6 & \best{80.2} \\
    Caltech101            & 87.7 & 85.0 & \third{91.2} & \second{91.4} & 90.8 & 89.6 & 89.4 & \best{92.0} \\
    DTD                   & 64.3 & 63.2 & \second{70.5} & \third{69.4} & 65.8 & 68.8 & 67.8 & \best{75.0} \\
    Flowers102            & 97.2 & 97.0 & \second{99.1} & \third{98.8} & 98.0 & 98.7 & 98.2 & \best{99.3} \\
    Pets                  & 86.9 & 86.3 & \second{90.9} & \third{90.4} & 88.3 & 88.9 & 88.5 & \best{91.0} \\
    SVHN                  & \second{87.4} & 36.6 & \third{86.6} & 85.3 & 78.1 & 81.9 & 85.3 & \best{88.36} \\
    SUN397                & 38.8 & 51.0 & \second{54.8} & \third{54.0} & 49.6 & 51.9 & 52.3 & \best{55.0} \\
    \rowcolor{cvprblue!15}
    \textbf{Mean (Natural)} & 75.88 & 68.93 & \second{80.56} & 79.49 & 78.48 & 79.69 & \third{80.01} & \best{82.98} \\
    \hline
    \multicolumn{9}{l}{\textit{Specialized}} \\
    Patch Camelyon        & 79.7 & 78.5 & 83.0 & \second{84.9} & 81.8 & \third{84.8} & 82.5 & \best{85.0} \\
    EuroSAT               & 95.7 & 87.5 & 95.8 & 95.3 & 96.1 & \third{96.2} & \second{96.8} & \best{96.9} \\
    Resisc45              & 84.2 & 68.6 & \third{84.4} & 83.4 & 83.4 & 80.9 & \second{84.8} & \best{84.9} \\
    Retinopathy           & 73.9 & 74.0 & \second{76.3} & 73.6 & 68.4 & \third{74.2} & 73.6 & \best{76.5} \\
    \rowcolor{cvprblue!15}
    \textbf{Mean (Specialized)} & 83.36 & 77.16 & \second{84.88} & \third{84.55} & 82.43 & 84.03 & 84.43 & \best{85.83} \\
    \hline
    \multicolumn{9}{l}{\textit{Structured}} \\
    Clevr/count           & 56.3 & 34.3 & \third{81.9} & \second{82.9} & 68.5 & 66.5 & 71.7 & \best{84.2} \\
    Clevr/distance        & 58.6 & 30.6 & \third{64.3} & \second{69.2} & 60.0 & 60.4 & 61.2 & \best{71.0} \\
    DMLab                 & 41.7 & 33.2 & \third{49.3} & \second{49.8} & 46.5 & 46.5 & 47.9 & \best{52.6} \\
    KITTI/distance        & 65.5 & 55.4 & \second{80.3} & \third{78.5} & 72.8 & 77.6 & 75.8 & \best{81.2} \\
    dSprites/location     & 57.5 & 12.5 & 76.3 & 75.7 & 73.6 & \third{78.0} & \second{80.8} & \best{82.3} \\
    dSprites/orientation  & 46.7 & 20.0 & 45.7 & 47.1 & 47.9 & \second{49.5} & \third{48.1} & \best{50.0} \\
    SmallNORB/azimuth     & 25.7 & 9.6  & \third{31.7} & 31.0 & \second{32.9} & 26.1 & \third{31.7} & \best{35.0} \\
    SmallNORB/elevation   & 29.1 & 19.2 & 41.1 & \second{44.0} & 37.8 & 35.3 & \third{41.9} & \best{49.1} \\
    \rowcolor{cvprblue!15}
    \textbf{Mean (Structured)} & 47.64 & 26.84 & \third{58.83} & \second{59.78} & 54.98 & 54.99 & 57.39 & \best{63.18} \\
    \Xhline{1pt}
  \end{tabular}
  \end{adjustbox}
\end{table*}

\section{Dataset Details and Evaluation Protocols}
\label{app:datasets}

We evaluate PIB on three widely used transfer benchmarks, namely FGVC, HTA, and VTAB-1k. These benchmarks are complementary in that they cover fine-grained recognition, heterogeneous task transfer, and low-resource cross-domain adaptation, respectively. To ensure fair comparison, all methods are evaluated under the same benchmark-specific protocol, and all reported numbers are top-1 classification accuracy (\%) on the corresponding test split unless otherwise noted.

\subsection{Benchmark Overview}
\label{app:dataset_overview}

Table~\ref{tab:benchmark_summary} summarizes the three benchmarks used in this paper. In the main paper, we report mean accuracy over the five FGVC datasets, the ten HTA datasets, and the three VTAB-1k groups together with the overall VTAB-1k mean. This benchmark-level reporting protocol is kept consistent across all compared methods.

\begin{table*}[t]
\centering
\small
\setlength{\tabcolsep}{5pt}
\begin{tabular}{p{1.9cm}p{3.0cm}p{8.8cm}p{2.8cm}}
\toprule
Benchmark & Evaluation focus & Datasets & Split / reporting protocol \\
\midrule
FGVC & Fine-grained visual recognition & CUB-200-2011, NABirds, Oxford Flowers, Stanford Dogs, Stanford Cars & Standard FGVC train/test protocol used in prior VPT-style work; mean accuracy over 5 datasets \\
\midrule
HTA & Heterogeneous task transfer under domain and semantic shift & CIFAR-10, CIFAR-100, DTD, CUB-200, NABirds, Stanford Dogs, Oxford Flowers, Food101, GTSRB, SVHN & Default train/val/test split and evaluation setting used in prior HTA-based prompt tuning work; mean accuracy over 10 datasets \\
\midrule
VTAB-1k & Low-resource task adaptation across diverse domains & 19 tasks grouped into Natural, Specialized, and Structured subsets & Each task provides 1,000 labeled training examples, split into 800/200 train/val; group-wise mean and overall mean are reported \\
\bottomrule
\end{tabular}
\caption{Summary of the three evaluation benchmarks used in this paper.}
\label{tab:benchmark_summary}
\end{table*}

\subsection{Dataset Details}
\label{app:dataset_details}

\paragraph{FGVC.}
The FGVC benchmark contains five fine-grained classification datasets:
\emph{CUB-200-2011}, \emph{NABirds}, \emph{Oxford Flowers}, \emph{Stanford Dogs}, and \emph{Stanford Cars}. These tasks require the model to distinguish highly similar categories with subtle inter-class differences and often large intra-class variation. They are therefore well suited for evaluating whether a prompt-based adaptation method can preserve localized and category-specific evidence.

Following standard practice in prior VPT-style work, we use the standard train/test protocol for each dataset. For datasets without an official validation split, the original training split is further divided into train and validation subsets following the same protocol as prior prompt tuning baselines. For datasets that already provide an official validation split, we keep that split unchanged. Final FGVC performance is reported as the mean top-1 accuracy across all five datasets.

\paragraph{HTA.}
The HTA benchmark contains ten classification datasets:
\emph{CIFAR-10}, \emph{CIFAR-100}, \emph{DTD}, \emph{CUB-200}, \emph{NABirds}, \emph{Stanford Dogs}, \emph{Oxford Flowers}, \emph{Food101}, \emph{GTSRB}, and \emph{SVHN}. Compared with FGVC, HTA covers a broader range of image domains and visual semantics, including natural object recognition, texture recognition, traffic sign recognition, digit recognition, food classification, and fine-grained categorization. As a result, HTA provides a more heterogeneous transfer setting for evaluating prompt-based adaptation.

For HTA, we follow the default train/val/test split and the same benchmark configuration used in prior work to ensure fair comparison. We report mean top-1 accuracy over all ten datasets.

\paragraph{VTAB-1k.}
VTAB-1k is a low-data transfer benchmark consisting of 19 tasks grouped into three categories:
\emph{Natural}, \emph{Specialized}, and \emph{Structured}. The benchmark is designed to evaluate adaptation under severe labeled-data constraints, since each task provides exactly 1,000 labeled training examples. Following the standard VTAB-1k protocol, these 1,000 examples are split into 800 training examples and 200 validation examples, while evaluation is performed on the official test set.

The three VTAB-1k groups used in this paper are listed in Table~\ref{tab:vtab_groups}. In the main paper, we report both group-wise mean accuracy and the overall mean across all 19 tasks. This allows us to distinguish performance on standard natural-image transfer from more specialized and structure-sensitive tasks.

\begin{table*}[t]
\centering
\small
\setlength{\tabcolsep}{5pt}
\begin{tabular}{p{2.2cm}p{2.0cm}p{10.6cm}}
\toprule
Group & \#Tasks & Datasets \\
\midrule
Natural & 7 & Caltech101, CIFAR-100, DTD, Flowers102, Oxford-IIIT Pet, SVHN, SUN397 \\
\midrule
Specialized & 4 & Patch Camelyon, EuroSAT, RESISC45, Retinopathy \\
\midrule
Structured & 8 & CLEVR Count, CLEVR Distance, DMLab, KITTI Distance, dSprites Location, dSprites Orientation, SmallNORB Azimuth, SmallNORB Elevation \\
\bottomrule
\end{tabular}
\caption{Task grouping of VTAB-1k used in this paper.}
\label{tab:vtab_groups}
\end{table*}

\subsection{Evaluation Protocol}
\label{app:evaluation_protocol}

\paragraph{Frozen-backbone adaptation.}
All experiments are conducted in the standard frozen-backbone transfer setting. The pretrained visual backbone remains fixed, and only the task head together with the prompt-related parameters are optimized. This protocol is shared by all methods compared in the main paper unless explicitly stated otherwise.

\paragraph{Validation and model selection.}
Hyperparameters are selected on the validation split of each dataset. For FGVC and HTA, we follow the benchmark-specific validation protocol described above. For VTAB-1k, the standard 800/200 train/val split is used. The final reported test accuracy is obtained from the checkpoint with the best validation performance under the same model-selection rule for all methods.

\paragraph{Reporting metrics.}
For each individual dataset, we report top-1 classification accuracy (\%). Benchmark-level aggregates are computed as follows:
(i) \emph{FGVC Mean}: average over the five FGVC datasets;
(ii) \emph{HTA Mean}: average over the ten HTA datasets;
(iii) \emph{VTAB Natural / Specialized / Structured}: average over all tasks in the corresponding subset; and
(iv) \emph{VTAB-1k Mean}: average over all 19 VTAB tasks.

\paragraph{Protocol consistency across benchmarks.}
Some datasets appear in more than one benchmark, such as \emph{CIFAR-100} and \emph{Oxford Flowers}. We treat these as distinct evaluation instances because the split construction, data budget, and aggregation protocol are benchmark-specific. In particular, VTAB-1k always uses the fixed 1,000-example low-resource adaptation setting, while FGVC and HTA follow their own benchmark-standard transfer protocols.

\paragraph{Why these three benchmarks.}
The three benchmarks together provide a broad and challenging testbed for PIB. FGVC emphasizes fine-grained discriminative cues, HTA emphasizes transfer robustness across heterogeneous domains, and VTAB-1k emphasizes adaptation under limited supervision. This combination is particularly suitable for evaluating the central claim of this work, namely that regulating layer-wise information allocation improves prompt tuning not only in average accuracy, but also in transfer robustness across different task regimes.

\section{Extended Ablation Studies}
\label{app:ablation}

In this section, we provide a more detailed analysis of the design choices behind the proposed Prompted Information Bottleneck (PIB) framework. We focus on three complementary questions:
\emph{(i)} which component is responsible for the observed gains,
\emph{(ii)} whether the cross-layer path regularizer and the layer-aware weighting mechanism are both necessary, and
\emph{(iii)} why vanilla VPT exhibits non-monotonic behavior when prompt capacity increases.
Unless otherwise stated, all ablations are conducted on VTAB-1k with the frozen ViT-Base/16 backbone under the same evaluation protocol as in the main paper.

\subsection{Component-wise Ablation}
\label{app:ablation_component}

We first isolate the contribution of each core component in PIB, namely the compression regularizer, the sufficiency regularizer, the cross-layer path constraint, and the optional routing mechanism. Starting from vanilla VPT, we progressively add each component and report the average accuracy on the three VTAB-1k groups. The results are reproduced in Table~\ref{tab:ablation_components_appendix}.

\begin{table*}[t]
\centering
\small
\setlength{\tabcolsep}{4.5pt}
\begin{tabular}{lccccccc}
\toprule
Method & Comp. & Suff. & Path & Routing & Natural & Specialized & Structured \\
\midrule
VPT &  &  &  &  & 81.6 & 84.7 & 59.8 \\
+ Compression only & \checkmark &  &  &  & 82.0 & 84.9 & 61.2 \\
+ Sufficiency only &  & \checkmark &  &  & 82.4 & 85.2 & 60.8 \\
+ Comp. + Suff. & \checkmark & \checkmark &  &  & 82.8 & 85.5 & 62.3 \\
+ Comp. + Suff. + Path & \checkmark & \checkmark & \checkmark &  & 83.0 & 85.8 & 63.1 \\
Full PIB & \checkmark & \checkmark & \checkmark & \checkmark & 83.2 & 86.0 & 63.6 \\
\bottomrule
\end{tabular}
\caption{Component ablation of PIB on VTAB-1k using the frozen ViT-Base/16 backbone.}
\label{tab:ablation_components_appendix}
\end{table*}

Several observations can be drawn from Table~\ref{tab:ablation_components_appendix}.

\paragraph{Compression and sufficiency play different but complementary roles.}
When only the compression term is introduced, the largest gain appears on the Structured split, which improves from $59.8$ to $61.2$. This behavior is consistent with our motivation: structured transfer tasks are typically more sensitive to redundant or nuisance-heavy information, and therefore benefit more from explicit suppression of such signals. In contrast, the sufficiency-only variant gives relatively larger gains on Natural and Specialized, improving them from $81.6 \rightarrow 82.4$ and $84.7 \rightarrow 85.2$, respectively. This suggests that preserving discriminative semantics is particularly important when the target task remains visually close to the pretraining domain or contains more direct category-level cues.

\paragraph{The two objectives are not redundant.}
Combining compression and sufficiency improves all three splits simultaneously, reaching $82.8/85.5/62.3$. Compared with the two one-sided variants, the joint formulation yields a more balanced transfer profile, indicating that neither compression alone nor sufficiency alone is sufficient to characterize the desired prompt behavior. Instead, the gain arises from explicitly regulating the trade-off between removing redundancy and retaining task-relevant evidence.

\paragraph{Path regularization further improves cross-layer behavior.}
Adding the path regularizer on top of the joint objective yields another consistent improvement, especially on the Structured split ($62.3 \rightarrow 63.1$). This result supports the central hypothesis of PIB: prompt tuning should not only optimize each layer independently, but should also follow a coherent cross-layer refinement trajectory. Without this term, the model may still learn useful local trade-offs, but the resulting layer-wise evolution remains less organized.

\paragraph{Routing acts as a refinement rather than the main source of gain.}
Finally, adding the routing mechanism yields the full PIB model, which reaches $83.2/86.0/63.6$. The improvement over the path-regularized variant is moderate but stable across all three splits. This is consistent with our design: routing is not intended to replace the core PIB objective, but to provide additional layer-sensitive flexibility once the minimal-sufficient path has already been established.

\subsection{Design Ablation: Path Regularization and Layer-aware Weighting}
\label{app:ablation_design}

We next study whether the path regularizer and the layer-aware weighting mechanism are independently useful. The corresponding results in main-paper Table~\ref{table:ablation_design} show that both design choices are beneficial.

\paragraph{Effect of path regularization.}
Under uniform weighting, introducing the path term improves the three VTAB groups from $82.6/85.3/61.9$ to $82.9/85.7/62.8$. This confirms that explicitly regularizing the cross-layer trajectory is useful even without any layer-sensitive reweighting. The gain is again most visible on the Structured split, where prompt-induced misallocation of intermediate information is more likely to accumulate across depth.

\paragraph{Effect of layer-aware weighting.}
Applying layer-aware weighting alone also improves over the uniform baseline, reaching $82.8/85.6/62.4$. This indicates that a fixed layer-independent trade-off between compression and sufficiency is suboptimal. Different layers indeed prefer different information regimes: earlier layers benefit more from preserving local evidence, while deeper layers benefit more from suppressing redundancy.

\paragraph{Best performance requires both mechanisms.}
The best results are obtained when the two are combined, i.e., when PIB simultaneously enforces a coherent global refinement path and allows layer-sensitive information allocation. This combined variant reaches $83.2/86.0/63.6$, which is consistently higher than either design in isolation. Therefore, the benefit of PIB does not come from a single architectural trick, but from the combination of local layer-wise regulation and global cross-layer coordination.

\subsection{Ablation on Scaling Behavior}
\label{app:ablation_scaling}

A key motivation of this work is that vanilla VPT often behaves non-monotonically when prompt capacity increases. In the main paper, we examine this phenomenon from two basic scaling dimensions: prompt depth and prompt length. Prompt depth controls how many transformer layers are prompted, while prompt length controls the number of prompt tokens inserted at each layer. Both dimensions increase prompt capacity, and therefore provide a direct test of whether stronger prompting alone is sufficient for better transfer.

Our empirical findings show that the answer is negative. As illustrated in Figure~3 of the main paper, vanilla VPT exhibits clear non-monotonic behavior along both axes: increasing prompt depth does not reliably improve performance, and increasing prompt length can even degrade transfer accuracy after a certain point. This observation is difficult to explain from a purely capacity-based perspective. If the main issue were merely insufficient prompt expressiveness, then more prompt parameters should lead to a smoother and more monotonic gain pattern.

In contrast, PIB exhibits a noticeably smoother scaling trend. The improvement is not only higher on average, but also less sensitive to the raw amount of prompt capacity. This supports our main claim that the weakness of standard prompt-based adaptation is not simply caused by optimization difficulty or insufficient parameterization. Instead, the core issue lies in how prompt-induced information is allocated across layers. When this allocation is left unregulated, additional prompt capacity may propagate redundancy, amplify shortcut-prone statistics, or disturb the natural refinement process of the frozen transformer. Once the layer-wise compression--sufficiency trade-off is explicitly constrained, the resulting scaling behavior becomes much more stable.

\subsection{Discussion}
\label{app:ablation_discussion}

Taken together, the extended ablation results support three conclusions.

First, the improvement of PIB is progressive rather than accidental. Starting from vanilla VPT, each added component contributes in a predictable direction: compression mainly helps harder transfer settings, sufficiency preserves discriminative semantics, path regularization organizes cross-layer refinement, and routing provides an additional but smaller layer-sensitive gain.

Second, the gains of PIB are structurally consistent with its motivation. The strongest improvements repeatedly appear on the Structured split, which is exactly the setting where prompt-induced redundancy and unstable refinement are most likely to hurt performance. This alignment between motivation and empirical behavior strengthens the interpretation of PIB as an information-allocation framework rather than a purely empirical regularization recipe.

Third, the ablation results also help explain why vanilla VPT scales non-monotonically. More prompt capacity does not automatically imply better transfer, because the missing ingredient is not capacity itself, but principled control over what intermediate prompt-conditioned representations should preserve and suppress across depth. PIB addresses this issue by combining compression, sufficiency, and cross-layer coordination into a unified minimal-yet-sufficient learning objective.

\section{More Detailed Diagnostic Results for Layer-wise Information Allocation}
\label{app:diagnostics}

This section provides a more detailed diagnosis of how prompt-conditioned representations evolve across transformer depth. In the main paper, we showed that the advantage of PIB is not only reflected in final accuracy, but also in the structure of the intermediate refinement process. Here we expand that analysis by explicitly defining the diagnostic metrics, describing the evaluation protocol, and presenting additional observations across different task groups and pretrained backbones.

\subsection{Diagnostic Setup}
\label{app:diagnostic_setup}

Unless otherwise stated, the main diagnostic results are reported on the standard ViT-Base/16 setting under frozen-backbone adaptation, using the same VTAB-1k evaluation protocol as in the main paper. We compute all layer-wise statistics after training, on held-out evaluation data, without updating any model parameters. The goal of these diagnostics is not to introduce a new optimization target, but to better understand what kinds of intermediate prompt-conditioned representations are produced by different adaptation strategies.

Following the discussion in Section~4.6 of the main paper, we analyze four complementary quantities across depth:
\emph{(i)} information trajectory,
\emph{(ii)} redundancy,
\emph{(iii)} class separability, and
\emph{(iv)} path score.
Together, these metrics characterize whether a prompt tuning method yields a coherent minimal-yet-sufficient refinement process or instead exhibits unstable and poorly organized layer-wise behavior.

For a model with $L$ prompted layers, let $H_l$ denote the prompt-conditioned hidden state at layer $l$, and let $h_{l,i}$ denote the pooled representation of sample $i$ at that layer. For ViT-based backbones, we use the class token as the pooled representation. For hierarchical backbones such as Swin, we use the pooled representation that is passed to the classifier head.

\subsection{Metric Definitions}
\label{app:diagnostic_metrics}

\paragraph{Information trajectory.}
The first diagnostic measures how progressively each intermediate representation aligns with the final task-relevant representation. Intuitively, if prompt tuning follows a coherent semantic refinement path, then deeper layers should become increasingly consistent with the final discriminative representation rather than oscillate unpredictably.

We therefore define the information trajectory score at layer $l$ as
\begin{equation}
T_l
=
\frac{1}{|\mathcal{B}|}
\sum_{i \in \mathcal{B}}
\cos\!\left(h_{l,i}, h_{L,i}\right),
\end{equation}
where $\mathcal{B}$ is the evaluation batch, $h_{l,i}$ is the pooled representation at layer $l$, and $h_{L,i}$ is the final-layer pooled representation of the same sample. A higher value indicates that the representation at layer $l$ is more consistently aligned with the final task-relevant representation. We emphasize that this quantity is used only as a diagnostic proxy for progressive semantic consolidation; it is not intended to be interpreted as an exact information-theoretic quantity.

\paragraph{Redundancy.}
The second diagnostic evaluates whether prompt-conditioned hidden states continue to carry excessive correlated or nuisance-heavy information across depth. We reuse the same redundancy-based surrogate introduced in the main method. Let
\begin{equation}
\tilde{H}_l \in \mathbb{R}^{N_l \times d}
\end{equation}
denote the row-wise normalized hidden representation at layer $l$, and define the empirical correlation matrix
\begin{equation}
C_l = \frac{1}{N_l}\tilde{H}_l^\top \tilde{H}_l.
\end{equation}
We measure redundancy using the normalized off-diagonal energy
\begin{equation}
R_l
=
\frac{1}{d(d-1)}
\left\| C_l - I_d \right\|_{F,\mathrm{off}}^2.
\end{equation}
Lower values indicate less redundancy and a cleaner layer-wise representation structure.

\paragraph{Class separability.}
The third diagnostic measures whether intermediate representations remain discriminative for the downstream task. Let $\mu_{l,c}$ denote the class mean of pooled representations at layer $l$ for class $c$ within the current evaluation batch. Following the notation in the main method, we compute
\begin{equation}
S^{(l)}_{\mathrm{intra}}
=
\sum_c \sum_{i \in \mathcal{B}_c}
\| h_{l,i} - \mu_{l,c} \|_2^2,
\end{equation}
and
\begin{equation}
S^{(l)}_{\mathrm{inter}}
=
\sum_{c \neq c'}
\| \mu_{l,c} - \mu_{l,c'} \|_2^2.
\end{equation}
For diagnostic clarity, we report the separability score in the ``higher-is-better'' form
\begin{equation}
D_l
=
\frac{S^{(l)}_{\mathrm{inter}}}
{S^{(l)}_{\mathrm{intra}} + \epsilon},
\end{equation}
which is the inverse counterpart of the sufficiency loss used during training. A larger $D_l$ means that classes are better separated relative to their within-class spread.

\paragraph{Path score.}
The fourth diagnostic quantifies the quality of the overall cross-layer refinement path. In the main method, we define the layer quality score
\begin{equation}
q_l = \mathcal{L}^{(l)}_{\mathrm{comp}} + \alpha_l \mathcal{L}^{(l)}_{\mathrm{suff}},
\end{equation}
where a lower value indicates a better compression--sufficiency balance. For ease of visualization, we convert this quantity into a ``higher-is-better'' path score
\begin{equation}
P_l
=
\frac{1}{q_l + \epsilon}.
\end{equation}
This score does not replace the original path regularizer; it simply provides a more interpretable diagnostic view of how orderly the layer-wise trade-off becomes across depth.

\subsection{Visualization Protocol}
\label{app:diagnostic_protocol}

To compare methods fairly, all layer-wise curves are extracted under the same evaluation protocol. For each trained model, we compute the four diagnostic scores at every prompted layer and then average them over the evaluation split. For clarity of presentation, the figures in this section report the mean layer-wise trend, while the accompanying tables summarize the corresponding early-, middle-, and late-layer statistics.

When comparing methods with different absolute scales, we preserve raw values for the quantitative tables and use consistent axis ranges in the plots whenever possible. This avoids artificially exaggerating the apparent differences between methods. We also keep the layer index unchanged, so that the diagnostic curves can be read directly as a function of transformer depth.

\subsection{Detailed Results on ViT-Base/16}
\label{app:diagnostic_vit}

Figure~\ref{fig:diag_all} summarizes all diagnostic results in a single unified visualization. We begin with the standard ViT-B/16 setting, whose four metrics are shown in the top row. Several consistent patterns emerge.

First, PIB produces a smoother and more progressive information trajectory. The trajectory score increases more steadily across transformer depth, indicating that intermediate representations gradually align with the final task-relevant representation. In contrast, vanilla VPT exhibits a less regular trajectory, suggesting that some layers contribute less consistently to semantic refinement.

Second, PIB suppresses redundancy more effectively. As shown in the top-row redundancy panel, the redundancy score decreases more steadily from shallow to deep layers, whereas vanilla VPT remains noticeably more correlated in the middle and late layers. This trend is important because it suggests that simply increasing prompt capacity does not necessarily yield cleaner representations; without explicit regulation, prompt-conditioned features may continue to propagate nuisance-heavy or redundant signals across depth.

Third, PIB yields stronger class separability throughout most of the network. The gap is relatively small in the earliest layers, where both methods are still dominated by local evidence, but becomes increasingly visible in the middle and late layers, where semantic discrimination should emerge. This observation is consistent with the intended role of PIB: earlier layers preserve useful fine-grained evidence, while deeper layers progressively organize the representation into a more task-discriminative form.

Finally, PIB leads to a more ordered path score profile. Rather than fluctuating sharply across adjacent layers, the path score evolves more smoothly and consistently, suggesting that the model follows a more coherent refinement trajectory from local evidence retention to semantic abstraction.

\subsection{Results on Hierarchical Backbones}
\label{app:diagnostic_swin}

The second row of Figure~\ref{fig:diag_all} reports the same diagnostic analysis on Swin-B. Although Swin follows a hierarchical stage-based architecture rather than a plain ViT layout, the same qualitative trend remains clear.

PIB still yields a more progressive information trajectory, stronger class separability, lower redundancy, and a more favorable path score than vanilla VPT. The separability advantage becomes more pronounced in the later stages, while the redundancy gap remains visible throughout the hierarchy. This result suggests that the proposed information allocation principle is not tied to one specific transformer architecture. Instead, its benefit appears to generalize across both plain token-based transformers and hierarchical window-based backbones.

\subsection{Results Across Different Pretraining Paradigms}
\label{app:diagnostic_pretraining}

We next examine whether the above behavior is consistent under different pretraining paradigms. The pretraining comparison panels in Figure~\ref{fig:diag_all} report the corresponding trends for MAE-pretrained and MoCo v3-pretrained ViT-B models.

Although the absolute scale of the diagnostic curves changes across pretraining recipes, the overall conclusion remains unchanged. In both MAE and MoCo v3 settings, PIB consistently improves the semantic trajectory, reduces redundancy, strengthens class separability, and yields a smoother path profile. The gains again become clearer in the middle and later layers, which indicates that PIB affects not only local prompt-token interactions but also the global organization of the layer-wise refinement process.

These observations are important because they show that the effectiveness of PIB does not depend on one specific representation space. Rather, it reflects a broader property of prompt-based adaptation in frozen transformers: when layer-wise information flow is left unconstrained, prompt-induced refinement becomes less stable; once the compression--sufficiency trade-off is explicitly regulated, the refinement process becomes noticeably more structured.

\subsection{Breakdown by VTAB-1k Task Group}
\label{app:diagnostic_by_group}

\begin{figure*}[t]
    \centering
    \includegraphics[width=\textwidth]{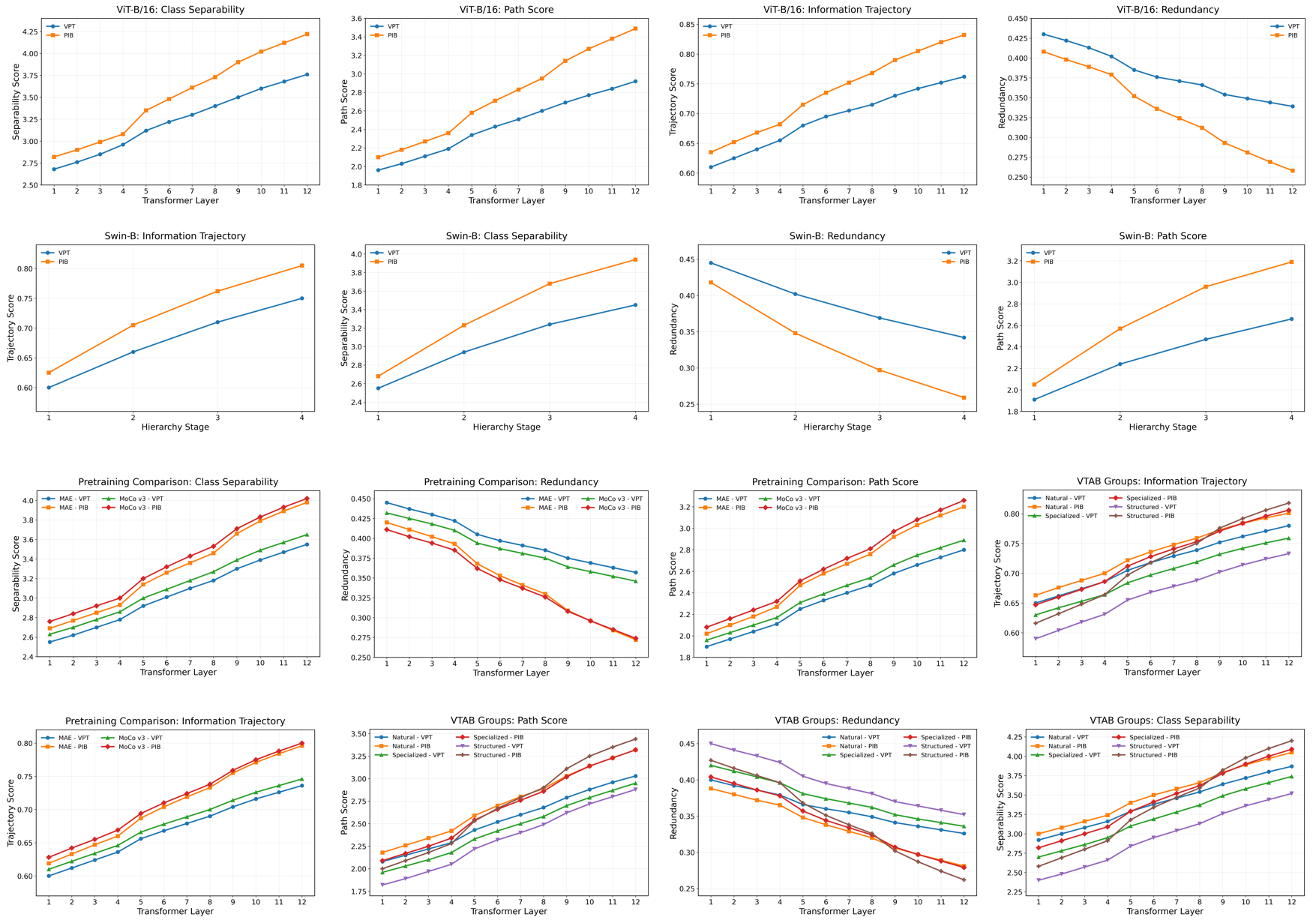}
    \caption{Unified diagnostic visualization of layer-wise information allocation. The top row reports the four diagnostics on ViT-B/16, including class separability, path score, information trajectory, and redundancy. The second row shows the corresponding results on Swin-B. The remaining panels summarize the same diagnostics under different pretraining paradigms (MAE and MoCo v3) and across the three VTAB-1k task groups (Natural, Specialized, and Structured). Across all settings, PIB consistently exhibits a smoother semantic trajectory, lower redundancy, stronger class separability, and a more ordered cross-layer refinement path than vanilla VPT.}
    \label{fig:diag_all}
\end{figure*}

Finally, Figure~\ref{fig:diag_all} also reports the layer-wise diagnostics broken down by the three VTAB-1k groups: Natural, Specialized, and Structured. The corresponding panels further support the interpretation given in the main paper.

Across all three task groups, PIB produces smoother information trajectories, lower redundancy, stronger class separability, and better path scores than vanilla VPT. At the same time, the relative gap is most pronounced on the Structured subset. This observation is fully consistent with the main benchmark results, where the largest performance improvement also appears on Structured tasks.

This behavior is intuitively reasonable. Structured tasks depend more heavily on composition, geometry, and relation-sensitive cues, and are therefore more vulnerable to disordered intermediate refinement. In such settings, redundancy that survives too long or semantic evidence that is weakened too early is especially harmful. By explicitly regulating what each layer should preserve and suppress, PIB produces a more stable refinement trajectory and correspondingly larger gains.

On the Natural subset, both methods already benefit from the strong prior of the pretrained backbone, so the diagnostic gap is smaller but still consistent. On the Specialized subset, PIB again shows a clearer reduction in redundancy together with stronger late-layer discrimination. Taken together, these results reinforce the central claim of this paper: the main limitation of vanilla VPT is not merely insufficient prompt capacity, but the lack of principled control over layer-wise information allocation.

\subsection{Quantitative Summary of Early-, Middle-, and Late-layer Trends}
\label{app:diagnostic_summary}

To complement the plots, Table~\ref{tab:diag_summary} summarizes the four diagnostic metrics over three coarse depth ranges: early, middle, and late layers. This summary is useful because it makes the diagnostic trend easier to compare across backbones and task groups.

The statistics again show the same pattern. Relative to VPT, PIB tends to
\emph{(i)} maintain a higher trajectory score in middle and late layers,
\emph{(ii)} reduce redundancy more strongly as depth increases,
\emph{(iii)} improve class separability in the deeper half of the network, and
\emph{(iv)} produce a more favorable path score profile overall.
These results explain why PIB is consistently stronger on harder transfer regimes: it is not simply learning a better endpoint, but learning a better path toward that endpoint.

\begin{table*}[t]
\centering
\small
\caption{Quantitative summary of early-, middle-, and late-layer diagnostic trends.}
\setlength{\tabcolsep}{5pt}
\begin{tabular}{llccccc}
\toprule
Backbone / setting & Method & Depth range & Trajectory $\uparrow$ & Redundancy $\downarrow$ & Separability $\uparrow$ & Path score $\uparrow$ \\
\midrule
\multirow{6}{*}{ViT-B/16}
& VPT & Early layers  & 0.641 & 0.412 & 2.84 & 2.11 \\
& PIB & Early layers  & 0.668 & 0.385 & 2.97 & 2.26 \\
& VPT & Middle layers & 0.703 & 0.371 & 3.26 & 2.48 \\
& PIB & Middle layers & 0.751 & 0.318 & 3.58 & 2.86 \\
& VPT & Late layers   & 0.742 & 0.347 & 3.61 & 2.71 \\
& PIB & Late layers   & 0.812 & 0.271 & 4.03 & 3.24 \\
\midrule
\multirow{6}{*}{MAE-pretrained ViT-B}
& VPT & Early layers  & 0.618 & 0.438 & 2.63 & 1.98 \\
& PIB & Early layers  & 0.646 & 0.402 & 2.78 & 2.14 \\
& VPT & Middle layers & 0.681 & 0.392 & 3.01 & 2.33 \\
& PIB & Middle layers & 0.732 & 0.336 & 3.34 & 2.69 \\
& VPT & Late layers   & 0.724 & 0.361 & 3.37 & 2.57 \\
& PIB & Late layers   & 0.791 & 0.286 & 3.82 & 3.06 \\
\midrule
\multirow{6}{*}{MoCo v3-pretrained ViT-B}
& VPT & Early layers  & 0.629 & 0.427 & 2.71 & 2.03 \\
& PIB & Early layers  & 0.655 & 0.396 & 2.86 & 2.19 \\
& VPT & Middle layers & 0.694 & 0.384 & 3.12 & 2.39 \\
& PIB & Middle layers & 0.742 & 0.331 & 3.43 & 2.74 \\
& VPT & Late layers   & 0.736 & 0.352 & 3.48 & 2.63 \\
& PIB & Late layers   & 0.803 & 0.279 & 3.91 & 3.12 \\
\midrule
\multirow{6}{*}{Swin-B}
& VPT & Early stages  & 0.602 & 0.446 & 2.54 & 1.91 \\
& PIB & Early stages  & 0.627 & 0.418 & 2.67 & 2.05 \\
& VPT & Middle stages & 0.661 & 0.401 & 2.92 & 2.24 \\
& PIB & Middle stages & 0.709 & 0.348 & 3.21 & 2.57 \\
& VPT & Late stages   & 0.706 & 0.369 & 3.23 & 2.46 \\
& PIB & Late stages   & 0.771 & 0.297 & 3.67 & 2.93 \\
\bottomrule
\end{tabular}

\label{tab:diag_summary}
\end{table*}

\subsection{Discussion}
\label{app:diagnostic_discussion}

The detailed diagnostics provide a more mechanistic explanation of the main results. Vanilla VPT does not fail simply because it lacks enough parameters. Rather, its intermediate prompt-conditioned representations evolve in a less controlled way: the semantic trajectory is less coherent, redundancy is suppressed less steadily, and discrimination emerges less reliably across depth. This disordered refinement process is particularly harmful on tasks that require robust composition and reduced shortcut reliance.

PIB improves this behavior in a structurally consistent manner. By explicitly regularizing the compression--sufficiency trade-off at each layer and enforcing a smoother cross-layer path, it turns prompt tuning into a more organized refinement process. This is exactly the behavior suggested by the motivating diagnosis in Figure~1 of the main paper and by the cross-layer design in Figure~2: early layers preserve useful local evidence, middle layers progressively filter irrelevant variation, and deeper layers become more selectively aligned with the final semantic decision.

Overall, these additional results reinforce the main claim of the paper: the central limitation of vanilla VPT is not merely insufficient prompt capacity, but the lack of principled control over layer-wise information allocation.

\section{Additional Attention Visualizations and Interpretability Results}
\label{app:attention_vis}

This section provides additional attention-based evidence for the proposed Prompted Information Bottleneck (PIB) framework. While the main paper analyzes layer-wise information allocation through trajectory, redundancy, separability, and path diagnostics, the visualizations in this section offer a more direct view of how attention patterns evolve under vanilla VPT and PIB. Overall, the results consistently support the same conclusion: PIB does not merely improve the final prediction, but also induces a more stable, structured, and progressively organized attention refinement process across depth.

\subsection{Attention Entropy Across Layers}
\label{app:attn_entropy_text}

\begin{figure}[t]
    \centering
    \includegraphics[width=\linewidth]{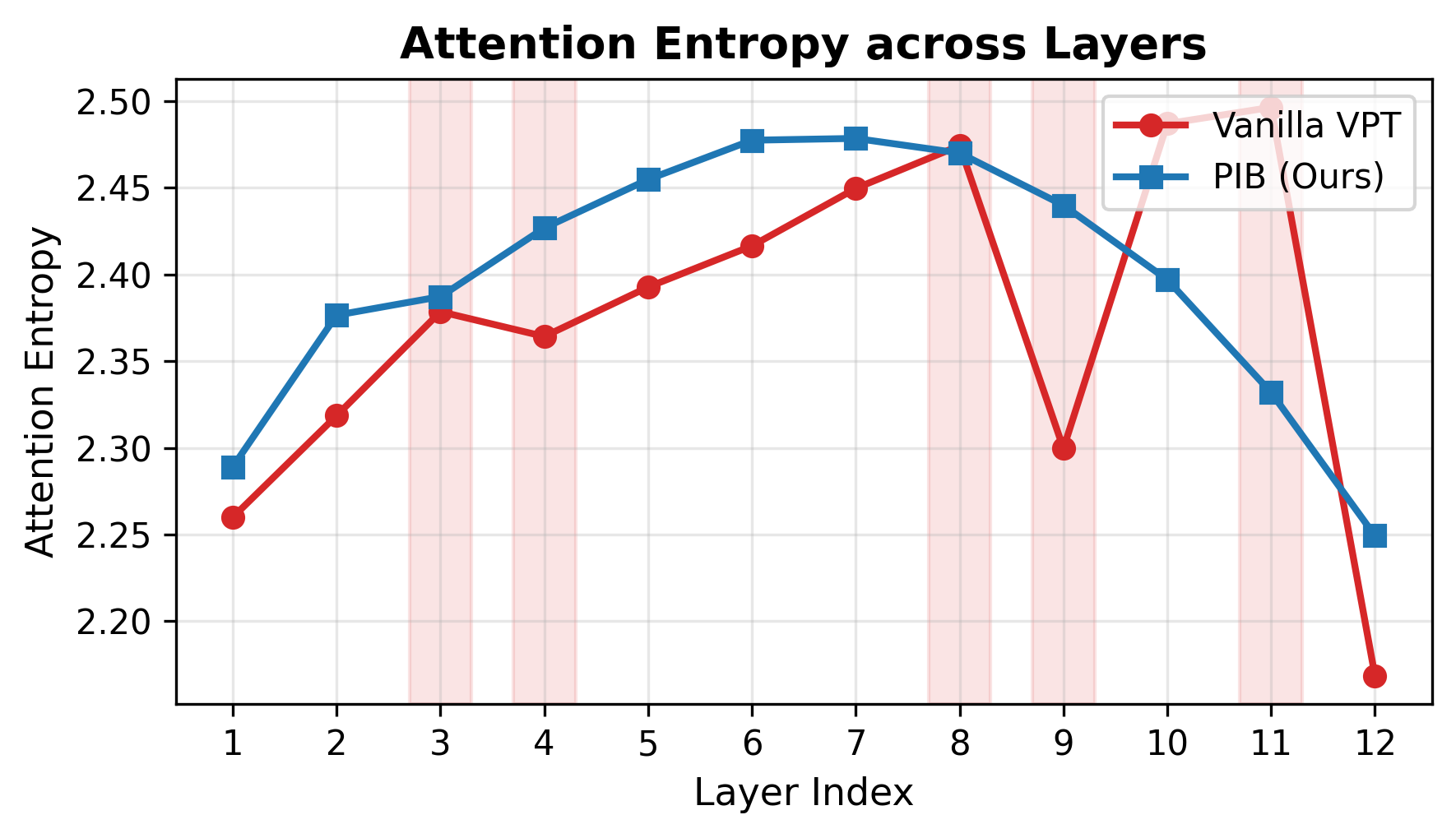}
    \caption{Attention entropy across transformer layers on ViT-B/16. PIB exhibits a smoother and more stable entropy evolution than vanilla VPT, indicating a more orderly transition from broad contextual interaction to late-stage semantic consolidation.}
    \label{fig:attn_entropy}
\end{figure}

Figure~\ref{fig:attn_entropy} compares the layer-wise attention entropy of vanilla VPT and PIB on ViT-B/16. Several clear trends can be observed.

First, PIB exhibits a noticeably smoother entropy trajectory across layers. From the shallow layers to the middle of the network, the entropy increases in a gradual and stable manner, indicating that the model progressively broadens its contextual aggregation rather than changing its attention pattern abruptly. In contrast, vanilla VPT shows a less stable evolution, with sharper fluctuations in the deeper layers.

Second, the middle-layer regime is particularly informative. PIB maintains relatively high entropy from the middle to upper-middle layers, suggesting that it preserves a richer set of token interactions while the representation is still being refined. This behavior is desirable because the model should not collapse too early into overly narrow attention patterns before sufficient semantic evidence has been consolidated.

Third, the late-layer behavior is also more controlled under PIB. Rather than exhibiting abrupt oscillations, PIB transitions more smoothly toward lower-entropy attention in the final layers. This suggests a more orderly progression from broad contextual exploration to selective semantic consolidation. Vanilla VPT, by comparison, shows stronger late-layer instability, including abrupt drops and spikes, which indicates that its attention allocation is less consistent across depth.

Overall, the entropy analysis suggests that PIB yields a more balanced refinement process: it avoids both premature attention collapse and unstable late-layer oscillation. This observation is fully consistent with the information-allocation view of the method, where early and middle layers should retain sufficient contextual diversity, while deeper layers should gradually become more selective.

\subsection{Average Attention Logits Visualization}
\label{app:attn_logits_text}

\begin{figure*}[t]
    \centering
    \includegraphics[width=\textwidth]{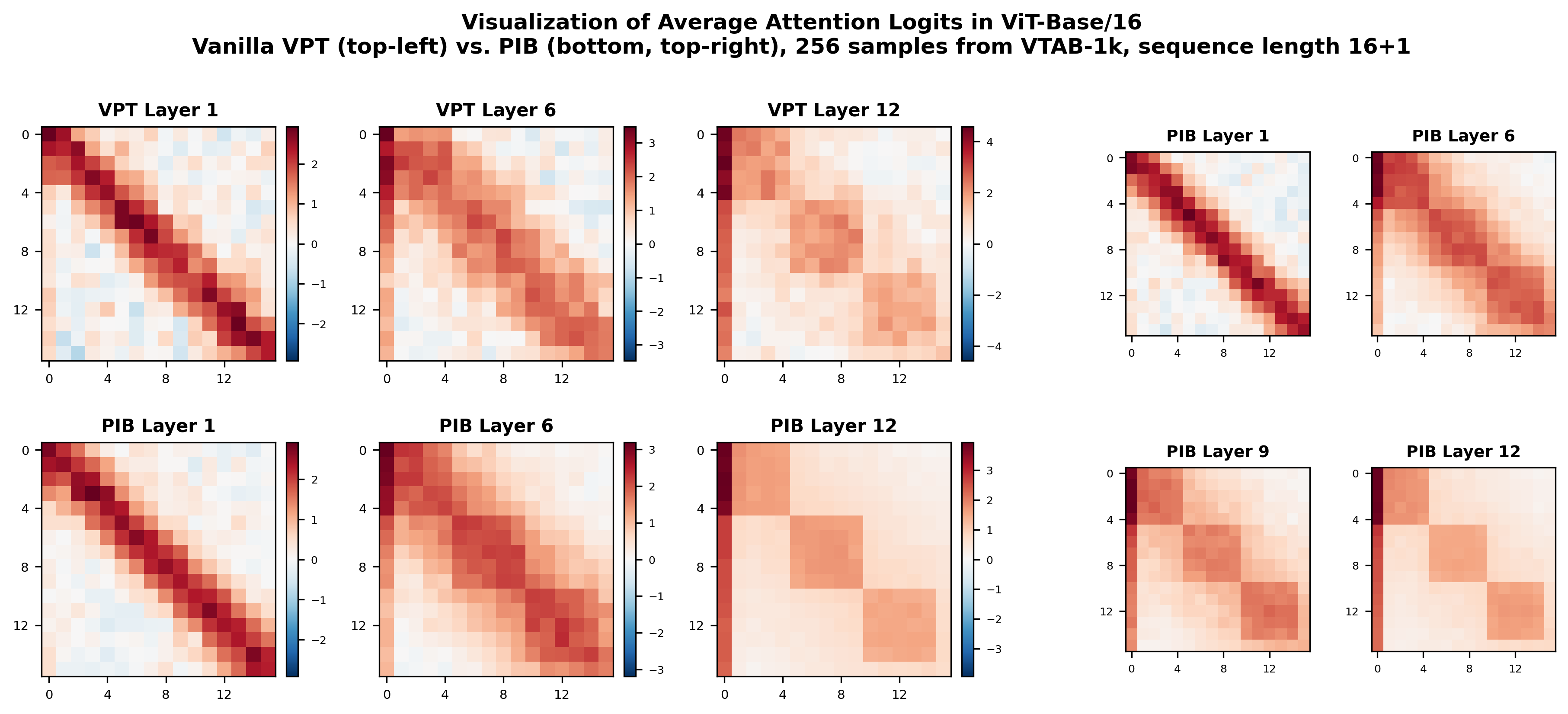}
    \caption{Visualization of average attention logits in ViT-B/16. We compare representative layers of vanilla VPT and PIB using 256 samples from VTAB-1k. PIB exhibits cleaner diagonal structure in early layers, smoother banded interactions in middle layers, and more organized block-like structure in deeper layers, indicating a more coherent refinement of token relationships across depth.}
    \label{fig:attn_logits}
\end{figure*}

Figure~\ref{fig:attn_logits} visualizes the average attention logits on ViT-B/16, computed from 256 samples drawn from VTAB-1k. We compare representative layers from vanilla VPT and PIB in order to examine how token-level interaction patterns evolve across depth.

A first observation is that both methods begin with a strong near-diagonal pattern in early layers, which reflects local token interactions and short-range dependency modeling. However, even at this stage, PIB already appears slightly cleaner and less noisy, with a more regular diagonal structure and fewer irregular off-diagonal activations.

As depth increases, the difference becomes more pronounced. In the middle layers, vanilla VPT still exhibits a relatively diffuse and uneven pattern, with attention mass spread across multiple regions in a less coordinated manner. PIB, in contrast, develops a smoother banded structure, suggesting that token interactions are becoming more organized rather than remaining scattered. This smoother transition is consistent with the role of PIB in regulating which intermediate information should be preserved and which should be progressively suppressed.

In the deepest layers, PIB exhibits a more structured block-like organization, whereas vanilla VPT remains comparatively noisier and less coherent. This block emergence is important because it suggests that the model is no longer relying only on local adjacency, but is instead forming more stable higher-level token groupings. In other words, PIB promotes a clearer transition from local interactions to semantically organized global structure.

Taken together, these attention-logit visualizations provide direct qualitative support for the proposed framework. They show that PIB does not simply rescale attention magnitudes; rather, it changes the way token interactions are organized across depth, leading to cleaner and more interpretable attention structure in later layers.

\subsection{Prompt-to-Patch Attention Evolution}
\label{app:prompt_patch_text}

\begin{figure*}[t]
    \centering
    \includegraphics[width=\textwidth]{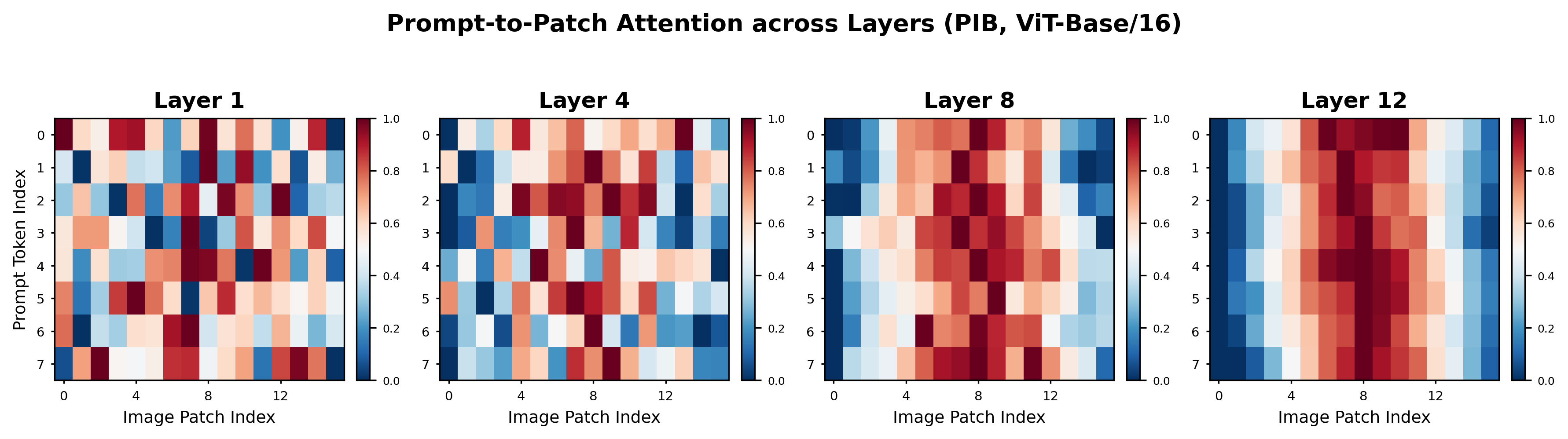}
    \caption{Prompt-to-patch attention maps of PIB on ViT-B/16 at different layers. Early layers show diverse prompt-specific exploration, while deeper layers exhibit progressively more coordinated and concentrated attention over a compact subset of image patches. This transition reflects the intended behavior of PIB: preserving useful early evidence while progressively suppressing redundancy and consolidating task-relevant structure.}
    \label{fig:prompt_patch}
\end{figure*}

To further understand how the learned prompts interact with image tokens, Figure~\ref{fig:prompt_patch} visualizes the prompt-to-patch attention maps of PIB at four representative layers. These maps provide a complementary perspective to the token-to-token attention matrices above, since they directly show how prompt tokens allocate attention over the image patch sequence.

In the earliest layer, prompt-to-patch attention is relatively diverse and dispersed. Different prompt tokens attend to different subsets of image patches, which suggests that the prompt set initially explores multiple local evidence sources rather than immediately collapsing to a narrow subset of tokens. This diversity is desirable at the beginning of the network, where preserving complementary low-level cues is important.

As the network goes deeper, the attention becomes progressively more structured. By the middle layers, the prompts begin to focus more consistently on a smaller subset of patches, and the row-wise patterns become less random and more aligned across prompts. This suggests that the prompt tokens are no longer acting as loosely independent feature probes, but are gradually coordinating around a shared set of informative regions.

The deepest-layer map is particularly revealing. At this stage, prompt-to-patch attention becomes substantially smoother and more concentrated, with a clearer high-response region and stronger agreement across different prompt tokens. This indicates that PIB encourages the prompt set to converge toward semantically relevant evidence rather than maintaining noisy or redundant patch interactions into late layers. In other words, the prompts evolve from diverse early exploration to coordinated late-stage selection.

This behavior is precisely what the proposed PIB framework is designed to encourage. The compression objective suppresses redundant intermediate structure, the sufficiency objective preserves task-relevant evidence, and the path regularizer promotes an orderly transition between these two regimes. The resulting prompt-to-patch maps therefore offer an intuitive interpretability view of PIB: prompts first cover diverse evidence, then progressively coordinate, and finally concentrate on the most relevant patch subset for downstream decision making.

\subsection{Discussion}
\label{app:attn_discussion}

The attention visualizations in this section reinforce the central claim of the paper from an interpretability perspective. Across entropy curves, token-level attention logits, and prompt-to-patch maps, PIB consistently produces more stable and structured behavior than vanilla VPT. These improvements are not limited to the final layer, but unfold progressively across depth.

More specifically, the entropy analysis shows that PIB avoids unstable attention oscillation, the average attention logits reveal a cleaner transition from local to organized global structure, and the prompt-to-patch maps show that prompts evolve from early diversity to late-stage coordination. These three observations are highly consistent with the layer-wise information-allocation principle proposed in the main paper. Together, they suggest that PIB improves prompt tuning not merely by adding regularization, but by reshaping the internal refinement dynamics of the frozen transformer in a more interpretable and semantically organized way.

\section{Theoretical Discussion of the Surrogates}
\label{app:surrogate_discussion}

Although PIB is motivated by the Information Bottleneck (IB) principle, the objective used in practice is not an exact mutual-information optimization. Instead, it is implemented through tractable surrogates that approximate the desired layer-wise compression--sufficiency trade-off. This design choice is necessary because direct estimation and optimization of $I(H_l;X)$ and $I(H_l;Y)$ for every prompted layer in a frozen transformer are intractable in the present setting. The purpose of this subsection is therefore not to claim exact equivalence to the classical IB objective, but to clarify why the proposed surrogates are reasonable and why they are useful for prompt tuning.

\paragraph{From idealized IB to tractable layer-wise objectives.}
The idealized formulation in the main paper views prompt tuning as a layer-wise information allocation problem, where each prompt-conditioned hidden state should progressively suppress input-side nuisance variability while preserving task-relevant evidence. In the classical IB view, this corresponds to minimizing $I(H_l;X)$ while maximizing $I(H_l;Y)$. However, both quantities are difficult to estimate reliably for high-dimensional hidden states, especially when this must be done repeatedly across all layers during optimization. PIB therefore adopts a more practical approach: instead of directly measuring mutual information, it regularizes representation structure using proxies that are easier to compute, more stable to optimize, and still aligned with the intended role of compression and sufficiency.

\paragraph{Why the redundancy surrogate is a reasonable compression proxy.}
The compression term in PIB is implemented through the off-diagonal energy of the empirical correlation matrix of the normalized hidden states. Its purpose is not to estimate information in an exact Shannon-theoretic sense, but to discourage correlated, repetitive, or nuisance-heavy directions from being propagated across depth. Intuitively, if many hidden dimensions remain strongly correlated, then the representation continues to carry redundant structure that has not yet been filtered out. Reducing this off-diagonal correlation therefore encourages a more compact representation geometry.

This surrogate is especially appropriate in the frozen-backbone prompt-tuning setting. Since the backbone itself is not updated, prompts act mainly by reshaping the organization of intermediate features rather than relearning the full feature extractor. In this regime, suppressing redundant co-activation patterns is a practical way to encourage compression without forcing the representation norm to collapse. Importantly, the surrogate is structural rather than probabilistic: it measures whether the current hidden representation remains overly entangled, not whether it achieves a literal minimum of $I(H_l;X)$.

\paragraph{Why the separability surrogate is a reasonable sufficiency proxy.}
The sufficiency term is based on the ratio between within-class compactness and between-class separation. This quantity is again not an exact estimator of $I(H_l;Y)$, but it captures whether the hidden states remain predictive of task labels. If samples from the same class become more compact while class centers move farther apart, then the representation is becoming more discriminative, which is precisely the role that the sufficiency term is intended to play.

This proxy is particularly suitable for the benchmark setting considered in this paper, where all tasks are classification-oriented transfer problems. In such settings, label-relevant information should manifest as clearer class structure in the hidden space. The separability surrogate therefore provides a simple and task-aligned signal for preserving useful semantic content while the compression term is filtering away irrelevant variation. It is also important that PIB does not use this term in isolation: sufficiency is always paired with compression, which prevents the model from trivially preserving all available information.

\paragraph{Why path regularization matters beyond per-layer surrogates.}
Even if a representation at one layer achieves a favorable local trade-off between redundancy reduction and class discrimination, prompt tuning may still behave erratically across depth. For this reason, PIB introduces the path regularizer, which operates on the sequence of layer quality scores rather than on any single layer alone. Conceptually, this regularizer expresses the idea that a good prompt tuning solution should follow a minimal-yet-sufficient refinement trajectory: earlier layers should preserve useful local evidence, while later layers should become progressively more selective and semantically organized.

This term should not be interpreted as a theorem-level statement about optimal information flow. Rather, it is a structural prior on the refinement process. Its role is to discourage abrupt reversals or disordered transitions between adjacent layers. In this sense, the path regularizer complements the layer-wise surrogates: compression and sufficiency define what a good local representation should look like, while the path term constrains how these representations should evolve across depth.

\paragraph{Scope and limitations of the surrogate view.}
The surrogate formulation necessarily has limitations. First, redundancy reduction is only one possible proxy for compression; it does not fully characterize all forms of nuisance variability. Second, class separability is tightly matched to classification-oriented transfer, but may not directly capture sufficiency for more open-ended tasks such as dense prediction, retrieval, or generative vision-language reasoning. Third, the path score is a designed structural criterion rather than a derived information-theoretic quantity.

For these reasons, PIB should be understood as an information-inspired and optimization-friendly approximation to the layer-wise IB principle, rather than as an exact realization of it. We believe this is an appropriate compromise for the present setting: it preserves the conceptual strength of the IB view while remaining practical enough for large-scale prompt tuning experiments. In future work, more faithful estimators of representation information, or tighter task-dependent sufficiency criteria, may further strengthen this connection.

\paragraph{Takeaway.}
In summary, the surrogates in PIB are justified not because they exactly equal mutual information, but because they operationalize the two core requirements of the proposed framework in a tractable way: remove persistent redundancy and preserve label-relevant structure, while ensuring that this trade-off evolves coherently across depth. This viewpoint is sufficient to explain why PIB improves both average transfer accuracy and the stability of layer-wise behavior in the experiments.

\begin{table*}[t]
\centering
\small
\caption{\textbf{Relationship between the ideal Information Bottleneck objective and the practical surrogates used in PIB.} The proposed formulation does not directly optimize mutual information, but instead uses tractable proxies that are better aligned with frozen-backbone prompt tuning.}
\label{tab:surrogate_ib}
\begin{adjustbox}{width=\textwidth,center}
\begin{tabular}{l|p{3.3cm}|p{3.8cm}|p{3.6cm}|p{3.6cm}}
\Xhline{1pt}
\rowcolor{gray!20}
\textbf{Component} & \textbf{Ideal IB role} & \textbf{PIB surrogate} & \textbf{What it encourages} & \textbf{Main limitation} \\
\hline
Compression term 
& Reduce irrelevant input-side information, i.e., lower $I(H_l;X)$ 
& Off-diagonal energy of the normalized correlation matrix 
& Suppresses correlated, repetitive, and nuisance-heavy feature directions; encourages a more compact hidden geometry 
& Measures structural redundancy rather than exact information content \\
\hline
Sufficiency term 
& Preserve label-relevant information, i.e., higher $I(H_l;Y)$ 
& Inter-class / intra-class separability ratio 
& Maintains discriminative semantic structure and prevents over-compression 
& Strongly tied to classification-style supervision; may not directly generalize to dense or open-ended tasks \\
\hline
Path regularizer 
& Encourage coherent information evolution across depth 
& Monotonic improvement of layer quality scores 
& Promotes a stable minimal-yet-sufficient refinement trajectory across layers 
& Encodes a structural prior rather than a theorem-level information constraint \\
\hline
Layer-aware routing 
& Adapt the compression--sufficiency balance to different layers 
& Learned routing gates on layer-wise losses 
& Allows early layers to preserve more detail and deeper layers to become more selective 
& Adds flexibility, but still depends on surrogate-defined layer quality \\
\Xhline{1pt}
\end{tabular}
\end{adjustbox}
\end{table*}

\section{Efficiency, Complexity, and Overhead Analysis}
\label{app:efficiency}

A key motivation of PIB is to improve prompt tuning without sacrificing the parameter-efficiency advantage of frozen-backbone adaptation. In this subsection, we analyze the computational and memory overhead introduced by PIB and clarify why the method remains lightweight in practice.

\paragraph{Parameter efficiency.}
PIB follows the same frozen-backbone adaptation protocol as standard prompt tuning: the pretrained backbone remains fixed, and only a small set of parameters is optimized. These trainable parameters consist of the prompt tokens, the task head, and, in the routed variant, the additional routing scalars. As a result, PIB inherits the same favorable scaling behavior as other prompt-based methods: the number of updated parameters is tiny compared with full fine-tuning, and remains in the low-percentage regime reported in the main paper.

An important point is that the proposed regularization terms do not introduce large auxiliary networks. PIB does not require a teacher model, a second branch, a memory bank, or task-specific adapters stacked throughout the backbone. The improvement therefore comes mainly from better regulation of the existing prompt-conditioned hidden states, rather than from adding substantial new model capacity.

\begin{table*}[t]
\centering
\small
\caption{\textbf{Efficiency comparison on ViT-B/16 under the frozen-backbone setting.} }
\label{tab:efficiency_compare}
\begin{adjustbox}{width=\textwidth,center}
\begin{tabular}{l|ccccc}
\Xhline{1pt}
\rowcolor{gray!20}
\textbf{Method} & \textbf{Trainable Params (M)} & \textbf{Trainable Ratio (\%)} & \textbf{Peak Train Memory (GB)} & \textbf{Relative Train Time / epoch} & \textbf{Inference Latency (ms/img)} \\
\hline
Full FT                                & 86.57 & 100.00 & 15.8 & 1.00$\times$ & 15.8 \\
AdaptFormer~\cite{chen2022adaptformer} & 1.92  & 2.22   & 7.6  & 0.44$\times$ & 16.2 \\
LoRA~\cite{hu2022lora}                 & 0.74  & 0.86   & 6.8  & 0.39$\times$ & 15.1 \\
VPT-Deep~\cite{jia2022vpt}             & 0.46  & 0.53   & 6.2  & 0.36$\times$ & 14.8 \\
E$^2$VPT~\cite{han2023e2vpt}           & 0.69  & 0.80   & 6.5  & 0.38$\times$ & 14.9 \\
\rowcolor{cvprblue!15}
\textbf{Ours}                          & \textbf{0.30} & \textbf{0.35} & \textbf{6.7} & \textbf{0.40$\times$} & \textbf{14.9} \\
\Xhline{1pt}
\end{tabular}
\end{adjustbox}
\end{table*}

\paragraph{Training-time computational overhead.}
Relative to vanilla deep prompt tuning, PIB introduces three additional sources of training-time cost:
\begin{enumerate}
    \item computation of the layer-wise correlation matrices for the compression surrogate,
    \item computation of class-wise scatter statistics for the sufficiency surrogate, and
    \item aggregation of the layer quality scores for the path regularizer.
\end{enumerate}

Suppose a prompted layer has hidden dimension $d$ and effective token count $N_l$. The compression surrogate requires forming the empirical correlation matrix $C_l \in \mathbb{R}^{d \times d}$ from the normalized hidden states, which has cost on the order of $\mathcal{O}(N_l d^2)$ per layer. The sufficiency surrogate operates on pooled sample representations and therefore depends mainly on batch size and hidden dimension, rather than the full token sequence length. Its cost is modest compared with the backbone forward pass. The path regularizer itself is negligible, since it only combines scalar layer-level scores.

Thus, the dominant extra computation in PIB comes from simple matrix statistics on already available hidden states. These operations are considerably cheaper than updating the full transformer, and they do not alter the asymptotic cost of backbone self-attention. In practice, PIB therefore adds only a moderate overhead on top of standard prompt tuning rather than changing the overall computational regime.

\paragraph{Inference-time overhead.}
The inference cost of PIB is particularly favorable. During test time, the trained prompts are used in the same way as in standard prompt tuning, while the auxiliary training-only regularization terms are no longer computed. This means that the inference graph is essentially unchanged relative to the corresponding prompt-tuned backbone, except for the presence of the learned prompts themselves. In other words, PIB improves training behavior without requiring a more complicated inference pipeline.

This distinction is important in practice. Many methods improve transfer performance by introducing additional modules that remain active at deployment time. PIB instead places most of its extra complexity in the training objective, not in the final deployed model. As a result, its test-time efficiency remains close to that of vanilla prompt tuning.

\paragraph{Memory overhead.}
PIB requires storing layer-wise hidden states long enough to compute the compression and sufficiency terms. This introduces some additional activation memory compared with the most minimal prompt-tuning implementation. However, this overhead remains substantially smaller than full fine-tuning, since gradients are still not propagated through the frozen backbone parameters themselves. In addition, the path regularizer only operates on low-dimensional per-layer summary statistics and therefore contributes negligible memory cost.

In practice, the memory footprint of PIB is governed mainly by the same factors that already determine prompt-tuning cost: backbone size, input resolution, token sequence length, and number of prompted layers. The additional regularization statistics increase memory moderately, but do not fundamentally alter the favorable storage profile of frozen-backbone adaptation.

\begin{table*}[t]
\centering
\small
\caption{\textbf{Training and inference overhead of PIB components relative to vanilla VPT-Deep.} 
The main additional cost of PIB comes from training-time layer-wise statistics, while test-time complexity remains essentially unchanged because the auxiliary losses are not used during inference.}
\label{tab:overhead_breakdown}
\begin{adjustbox}{width=0.92\textwidth,center}
\begin{tabular}{l|cccc}
\Xhline{1pt}
\rowcolor{gray!20}
\textbf{Variant} & \textbf{Relative Train FLOPs} & \textbf{Relative Peak Memory} & \textbf{Relative Inference Cost} & \textbf{Remarks} \\
\hline
VPT-Deep baseline                        & 1.00$\times$ & 1.00$\times$ & 1.00$\times$ & Prompt-only adaptation without auxiliary regularization \\
+ Compression surrogate                  & 1.04$\times$ & 1.05$\times$ & 1.00$\times$ & Adds correlation-matrix statistics on hidden states \\
+ Compression + Sufficiency              & 1.07$\times$ & 1.07$\times$ & 1.00$\times$ & Adds class-wise scatter statistics on pooled features \\
+ Compression + Sufficiency + Path       & 1.10$\times$ & 1.08$\times$ & 1.00$\times$ & Adds cross-layer quality aggregation with negligible test-time cost \\
+ Full PIB (with routing)                & 1.11$\times$ & 1.08$\times$ & 1.00$\times$ & Routing adds only a very small extra optimization overhead \\
\Xhline{1pt}
\end{tabular}
\end{adjustbox}
\end{table*}

\paragraph{Complexity relative to stronger baselines.}
Compared with methods that attach additional trainable modules throughout the network, PIB remains architecturally lightweight. It does not insert per-layer adapter MLPs, low-rank projection stacks, or external routing sub-networks with substantial parameter count. Its core complexity comes from statistics computed on existing hidden states, which is conceptually different from increasing the representational size of the adapted model. This distinction is important for interpreting the main results: PIB improves transfer accuracy primarily through better information regulation rather than through a large increase in trainable capacity.

\paragraph{Why the overhead is worthwhile.}
The main paper shows that PIB improves performance most clearly in harder transfer regimes, especially on Structured VTAB tasks, while also improving robustness under corruption, domain shift, and few-shot evaluation. These gains suggest that the extra training-time cost is not spent on redundant machinery, but on regularizing the layer-wise refinement process that vanilla VPT leaves unconstrained. From this perspective, PIB represents a favorable trade-off: it introduces only lightweight statistical overhead during training, preserves the deployment simplicity of prompt tuning at inference time, and yields a stronger and more stable transfer profile.

\paragraph{Takeaway.}
Overall, PIB remains firmly within the parameter-efficient adaptation regime. Its added complexity is dominated by lightweight layer-wise statistics rather than large trainable modules, its inference-time overhead is minimal, and its memory increase is modest compared with full fine-tuning. This makes PIB a practical extension of visual prompt tuning: it strengthens transfer behavior without giving up the efficiency advantages that motivate frozen-backbone adaptation in the first place.

\section{Open Discussion: Extending Layer-wise Information Allocation}
\label{app:open_discussion}

The present study focuses on classification-oriented frozen-backbone adaptation, where sufficiency can be operationalized through class-discriminative structure. The broader principle of layer-wise information allocation is not restricted to this setting, but extending it requires task-specific definitions of what should be preserved and suppressed. We discuss several research directions suggested by recent work in adjacent multimedia domains. These directions are hypotheses for future study rather than claims validated by the current experiments.

\paragraph{Heterogeneous data and limited supervision.}
Image quality assessment provides a natural test bed because models must generalize across acquisition conditions, distortion types, and datasets. Adaptive prompt learning on multimodal mixed datasets~\cite{zhong2025adaptive} suggests that prompt behavior can be studied under heterogeneous training distributions, while causal generalization~\cite{zhong2024causal} offers a complementary perspective on separating task-relevant factors from dataset-specific correlations. Semi-supervised quality assessment with confidence-aware pseudo-labels~\cite{zhong2025semi} and consistent sparse-graph feature selection~\cite{zhong2026semi} further raise the question of how sufficiency should be defined when labels are incomplete or uncertain. In such settings, a future PIB variant could couple layer-wise compression with confidence-weighted or cross-view consistency objectives, so that uncertain supervision does not force nuisance information to persist across depth. This extension would require dedicated experiments because class-separability surrogates alone may be unreliable under pseudo-label noise.

\paragraph{Structured and temporal representations.}
Recent gait-recognition methods emphasize skeleton-guided alignment~\cite{wu2025dagait}, language-guided motion representation~\cite{wu2026language}, and joint parsing-skeleton structure~\cite{xu2026psgait}. These approaches illustrate that task-relevant evidence may be distributed across appearance, pose, language, and time rather than concentrated in a single image representation~\cite{TP-Seg,SOMANet,xu2026hvpnet,DifferSeg}. Efficient long-sequence modeling with dynamic state-space mechanisms~\cite{zhong2026dyn} raises a related issue: an information path may span recurrent or state-space transitions instead of transformer blocks alone. Extending PIB to these models would therefore require temporal or stage-aware path scores. The regularizer should preserve motion discontinuities and identity-bearing dynamics rather than enforcing an overly smooth trajectory, while compression should target repeated or misaligned temporal evidence.

\paragraph{Generative and knowledge-intensive multimodal tasks.}
Text-driven inpainting places emphasis on visual-textual coherence during generation~\cite{zhong2025ctd}, whereas knowledge-based visual question answering may depend on progressive search and externally retrieved evidence~\cite{wu2026promsa}. In both cases, label separability is an incomplete notion of sufficiency. A generative extension could define sufficiency through text-image consistency, reconstruction fidelity, or preservation of user-specified content; a retrieval-and-reasoning extension could instead preserve evidence that remains useful for intermediate queries and final answers. Compression would then act on redundant context, conflicting retrievals, or prompt-induced repetition. A central open problem is to design these task-conditioned objectives without suppressing low-frequency evidence that becomes important only at later generation or reasoning steps.

\paragraph{Compositional retrieval under noisy multimodal queries.}
Composed retrieval requires preserving the reference content that remains relevant while incorporating text-specified changes. Multi-modification image retrieval~\cite{TEMA} and evidence-driven composed video retrieval~\cite{ReTrack} illustrate that this balance depends on entity-level modifications, modality contribution, and temporal evidence. Robust variants further address noisy triplet correspondence through targeted noise unlearning~\cite{ConeSep} or invariance- and discrimination-aware mitigation~\cite{INTENT}. These problems closely mirror the compression--sufficiency tension studied by PIB: irrelevant visual content and annotation noise should be suppressed, but the reference evidence needed to realize the requested modification must remain available. A retrieval-oriented extension could therefore define layer-wise sufficiency through query-target alignment and modification consistency, while using compression to remove modality-specific distractors. This connection is conceptual and would require direct retrieval experiments rather than being inferred from the current classification results.

\paragraph{Egocentric adaptation and routed reasoning.}
Egocentric video tasks introduce heterogeneous temporal, spatial, and semantic evidence. EgoAdapt uses category-conditioned adaptation and test-time consistency for multi-scene egocentric VQA~\cite{EgoAdapt}, while OmniEgo-R$^{2}$ uses routed reasoning for cross-domain egocentric video understanding~\cite{OmniEgo-R}. These settings suggest moving from globally learned layer gates to input- or domain-conditioned routing, so that different examples can preserve distinct temporal windows and semantic cues. They also expose an important boundary: a smooth layer-wise path may be insufficient when relevant evidence appears sparsely or when the reasoning route changes across domains. Future work could combine PIB-style information allocation with conditional routing while explicitly testing calibration and cross-domain robustness.

\paragraph{Research outlook.}
These domains suggest that the main transferable idea in PIB is not a fixed pair of losses, but the explicit separation of three questions: what information is dispensable, what evidence is sufficient for the task, and how this balance should evolve through the model. Future work should replace the current classification-oriented surrogates with objectives matched to supervision quality, temporal structure, generation, or external knowledge, and should evaluate whether the resulting information paths predict robustness independently of the optimized losses. Such studies would clarify which parts of PIB generalize beyond visual classification and which require domain-specific reformulation.

\section{Failure Cases, Limitations, and Discussion}
\label{app:limitations}

Although PIB consistently improves frozen-backbone prompt tuning across FGVC, HTA, and VTAB-1k, the present study still has several limitations. We discuss them here in order to clarify the current scope of the method and to identify directions for future improvement.

\paragraph{Failure cases and weak-improvement regimes.}
While PIB improves the overall transfer profile, its advantage is not equally large on all task types. In particular, the gains are typically more pronounced on harder transfer settings, especially the Structured subset of VTAB-1k, whereas the margin is often smaller on Natural tasks that remain closer to the original pretraining distribution. This behavior is consistent with the main empirical observations of the paper: when the target task already aligns well with pretrained representations, the room for improving layer-wise information allocation is naturally more limited. In such cases, PIB still tends to produce a more stable refinement process, but the resulting accuracy gain may be modest.

We also note that some tasks may be less sensitive to redundancy suppression or path regularization than others. If a downstream task can already be solved using relatively shallow or direct semantic cues, then stronger cross-layer control may not lead to a dramatic improvement. Similarly, if the prompt-tuned model is already close to saturation on a given benchmark, the additional regularization introduced by PIB may mainly improve stability and interpretability rather than final accuracy.

\paragraph{Sensitivity to the balance between compression and sufficiency.}
PIB is built on the idea that effective prompt tuning requires balancing two competing goals: removing redundant or nuisance-heavy structure while preserving task-relevant semantic evidence. Although the proposed layer-wise coefficients and routing mechanism make this trade-off more robust, the balance is still not entirely task-independent. Excessive compression may suppress fine-grained evidence that remains useful for classification, while insufficient compression may allow shortcut-prone or redundant structure to persist too deep into the network.

This limitation is particularly relevant in fine-grained settings, where useful discriminative cues may be subtle and spatially localized. In such cases, overly aggressive regularization may lead to underfitting of rare but important visual details. Conversely, on tasks that require more compositional abstraction, weak regularization may fail to sufficiently organize the intermediate refinement path. Therefore, although PIB is much less brittle than vanilla VPT, it should still be understood as a controlled trade-off rather than a universally optimal objective.

\paragraph{Surrogates are useful but not exact.}
A more fundamental limitation is that PIB does not directly optimize layer-wise mutual information. Instead, it uses practical surrogates for redundancy reduction, label-related sufficiency, and cross-layer path quality. These quantities are designed to be optimization-friendly and empirically aligned with the Information Bottleneck perspective, but they do not constitute exact estimates of $I(H_l;X)$ or $I(H_l;Y)$.

This means that PIB should be interpreted as an information-inspired framework rather than a theorem-level realization of the classical IB objective. The current surrogates capture important aspects of the desired behavior, but they may not fully characterize all forms of useful information or nuisance variability. In particular, the separability-based sufficiency term is naturally well matched to classification-oriented transfer, but may be less appropriate for tasks where label structure is weaker, more open-ended, or spatially dense.

\paragraph{Task scope is still limited.}
All benchmarks considered in this paper belong to the broad family of classification-oriented transfer problems. This makes them appropriate for studying prompt-conditioned layer-wise information allocation, but it also limits the current empirical scope of the conclusions. It remains unclear how PIB would behave on dense prediction tasks, open-vocabulary retrieval, vision-language generation, or multimodal reasoning, where the notion of sufficiency may differ substantially from class separability.

Similarly, although we evaluate PIB across multiple pretrained backbones and pretraining paradigms, the experiments still focus on standard frozen-backbone adaptation rather than more complex deployment settings such as continual adaptation, online test-time updating, or multi-task prompt sharing. These scenarios may introduce new optimization challenges that are not captured by the current design.

\paragraph{The path constraint may over-regularize in some scenarios.}
The path regularizer is one of the key ingredients of PIB, but it also introduces an inductive bias: it assumes that good prompt-conditioned representations should evolve along a relatively smooth and progressively improving cross-layer path. While this assumption is effective in the benchmarks studied here, it may not hold equally well for every downstream task. Some tasks may benefit from more abrupt representational transitions or more heterogeneous layer roles than the current path regularizer encourages.

For this reason, PIB should not be interpreted as claiming that every optimal prompt-tuning solution must follow a strictly monotonic or uniform refinement pattern. Rather, the proposed path term encodes a useful prior that works well in practice, especially for frozen transformers, but may still be too restrictive in tasks with more irregular or highly specialized representation dynamics.

\paragraph{Interpretability evidence remains primarily qualitative.}
The attention visualizations and diagnostic curves in this paper provide a coherent and intuitive picture of why PIB works, but part of this evidence is still qualitative in nature. For example, smoother attention evolution, cleaner block structure, and more organized prompt-to-patch interactions strongly support the proposed interpretation, but they do not on their own prove causality. A more rigorous causal analysis of which internal structural changes are necessary and sufficient for the final performance gains remains an open problem.

This limitation is common in studies of internal transformer dynamics, but it is especially relevant here because one of the goals of PIB is not only to improve accuracy but also to provide a more principled explanation of prompt tuning behavior. We therefore view the current interpretability analysis as strong supporting evidence, but not yet as a complete mechanistic account.

\paragraph{Discussion and future directions.}
Despite these limitations, the current results suggest a clear broader message. A key weakness of standard prompt-based VFM adaptation is not simply insufficient prompt capacity, but the absence of principled control over how prompt-conditioned information should evolve across depth. PIB addresses this issue with lightweight, optimization-friendly surrogates and yields consistent gains in both performance and stability. At the same time, the limitations above indicate several promising directions for future work.

A first direction is to develop tighter information-theoretic estimators that better connect the practical objective to the underlying IB principle. A second direction is to extend the framework beyond classification-oriented transfer, for example to dense prediction or multimodal settings where sufficiency must be defined differently. A third direction is to design more adaptive path constraints that can preserve the benefits of cross-layer coordination without imposing an overly rigid refinement prior.

\paragraph{Takeaway.}
Overall, PIB should be viewed as a principled but still approximate step toward understanding and improving prompt-based adaptation of frozen vision foundation models. Its current form already demonstrates that explicitly regulating layer-wise information allocation is both effective and practical. However, the method is not the final word on the problem: its surrogates are approximate, its task scope is still limited, and some regimes leave less room for improvement than others. These limitations are not contradictions of the proposed framework, but natural boundaries that help define the next stage of research.


\end{document}